\pdfoutput=1

\documentclass[11pt]{article}

\usepackage{arabtex} 
\usepackage{utf8}
\usepackage[utf8]{inputenc}
\usepackage{EMNLP2023}

\usepackage[shortlabels]{enumitem}
\usepackage{times}
\usepackage{color,soul} 
\usepackage{booktabs}
\usepackage{latexsym}
\usepackage{hyperref}
\usepackage{multirow}
\usepackage[LAE,T1]{fontenc}

\usepackage[utf8]{inputenc} 
\usepackage[LAE,T1]{fontenc}
\usepackage[arabic,french,english]{babel}
\usepackage[raggedrightboxes]{ragged2e}
\usepackage{todonotes}
\usepackage{lscape}
\usepackage{natbib}
\usepackage{longtable}
\usepackage{enumitem}
\usepackage{float}
\usepackage{makecell}
\usepackage{inconsolata}
\usepackage{arydshln}
\usepackage{soul}

\usepackage{graphicx}
\usepackage{scrtime}
\usepackage[dont-mess-around]{fnpct}
\usepackage[utf8]{inputenc}
\usepackage[ arabic , english]{babel}
\usepackage{colortbl}
\usepackage{etoolbox}
\usepackage{dirtytalk}
\usepackage{array}
\usepackage{pmboxdraw}
\usepackage{bbding}
\usepackage{changepage}
\usepackage{natbib}
\usepackage{times}
\usepackage{latexsym}
\usepackage[utf8]{inputenc}
\usepackage{arabtex}
\usepackage{colortbl}

\usepackage[locale=FR]{siunitx}
\usepackage{booktabs}
\usepackage[T1]{fontenc}
\usepackage{textcomp}

\usepackage{appendix}

\usepackage{geometry}
\usepackage{multirow}
\usepackage{tipa}  

 \usepackage{array,multirow,graphicx}
 \usepackage{float}
 \usepackage{colortbl}

\usepackage{multirow}
\usepackage{array}
\usepackage{rotating}
\usepackage{arabtex} 
\usepackage{graphicx} 
\usepackage{lscape} 
\usepackage{geometry} 

\usepackage{rotating}

\definecolor{olivegreen}{RGB}{107,142,35}
\definecolor{lightolivegreen}{RGB}{157,192,105}

\definecolor{blue}{RGB}{0,0,255}
\definecolor{lightblue}{RGB}{0,216,230}

\usepackage{microtype}

\usepackage{rotating}
\usepackage{amsmath}

\usepackage{inconsolata}

\usepackage{graphicx}
\usepackage{subcaption}
\usepackage{tabularx}
\usepackage{framed}

\setcode{utf8} 
\usepackage[T1]{fontenc}

\usepackage[utf8]{inputenc}

\usepackage{microtype}

\usepackage[utf8]{inputenc}
\usepackage{tikzsymbols}
\usepackage{microtype}

\def\r{\color{red}}
\definecolor{blush}{rgb}{0.87, 0.36, 0.51}

\definecolor{dgreen}{rgb}{0.0, 0.5, 0.0}
\def\g{\color{dgreen}}
\def\n{\color{black}}

\title{
\raisebox{-2.1ex}{\protect\includegraphics[height=3.8\fontcharht\font`\B]{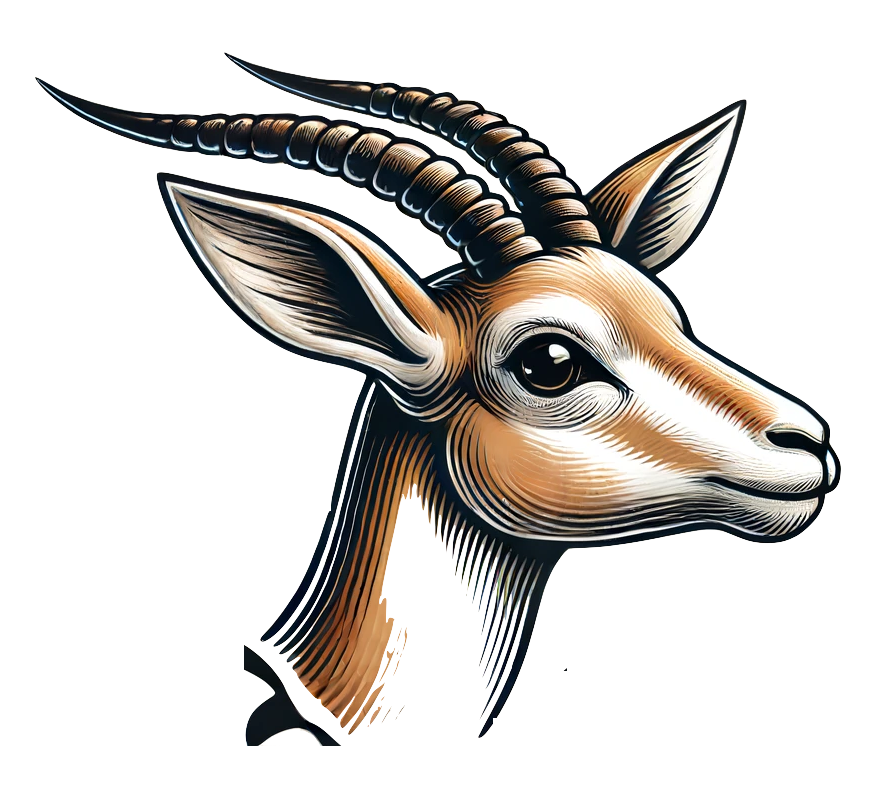}}
 \Large \bfseries Gazelle: An Instruction Dataset for Arabic Writing Assistance}

\author{\normalsize Samar M. Magdy$^{\xi}$~ Fakhraddin Alwajih$^{\lambda}$\thanks{~~Equal contribution}~ Sang Yun Kwon$^{\lambda}$\footnotemark[1] \\ 
~\normalsize\textbf{Reem Abdel-Salam}$^{\phi}$ ~\normalsize\textbf{Muhammad Abdul-Mageed}$^{\lambda,\xi,\gamma}$~  
 \\
\normalsize $^{\lambda}$The University of British Columbia~~~~ $^{\xi}$MBZUAI \\
\normalsize  $^{\gamma}$Invertible AI ~~~~ $^{\phi}$Cairo University\\
   \texttt{\normalsize {samar.magdy}@mbzuai.ac.ae, {muhammad.mageed}@ubc.ca}
}

\setcode{utf8} 
\setarab 
\begin{document}
\maketitle
\begin{abstract}
Writing has long been considered a hallmark of human intelligence and remains a pinnacle task for artificial intelligence (AI) due to the intricate cognitive processes involved. Recently, rapid advancements in generative AI, particularly through the development of Large Language Models (LLMs), have significantly transformed the landscape of writing assistance. However, underrepresented languages like Arabic encounter significant challenges in the development of advanced AI writing tools, largely due to the limited availability of data. This scarcity constrains the training of effective models, impeding the creation of sophisticated writing assistance technologies. To address these issues, we present \textit{Gazelle}, a comprehensive dataset for Arabic writing assistance. In addition, we offer an evaluation framework designed to enhance Arabic writing assistance tools. Our human evaluation of leading LLMs, including GPT-$4$, GPT-$4o$, Cohere Command R+, and Gemini $1.5$ Pro, highlights their respective strengths and limitations in addressing the challenges of Arabic writing. Our findings underscore the need for continuous model training and dataset enrichment to manage the complexities of Arabic language processing, paving the way for more effective AI-powered Arabic writing tools.
\end{abstract}

\section{Introduction}\label{sec:intro}
Writing is an essential skill for both individuals and professionals, requiring a long-term commitment to mastery that can span years of continuous learning and refinement~\cite{Flower, du-etal-2022-understanding-iterative}. The distinct cognitive processes involved in writing underscore the significant importance and complexity of writing as a task. Consequently, in the fields of AI and natural language generation (NLG), writing is considered a hallmark of human intelligence, and researchers have long viewed it as a pinnacle task for their studies~\cite{ippolito2022creative,book_woo,article_chens}.
\begin{figure}[t]
  \centering
  \includegraphics[width=\linewidth]{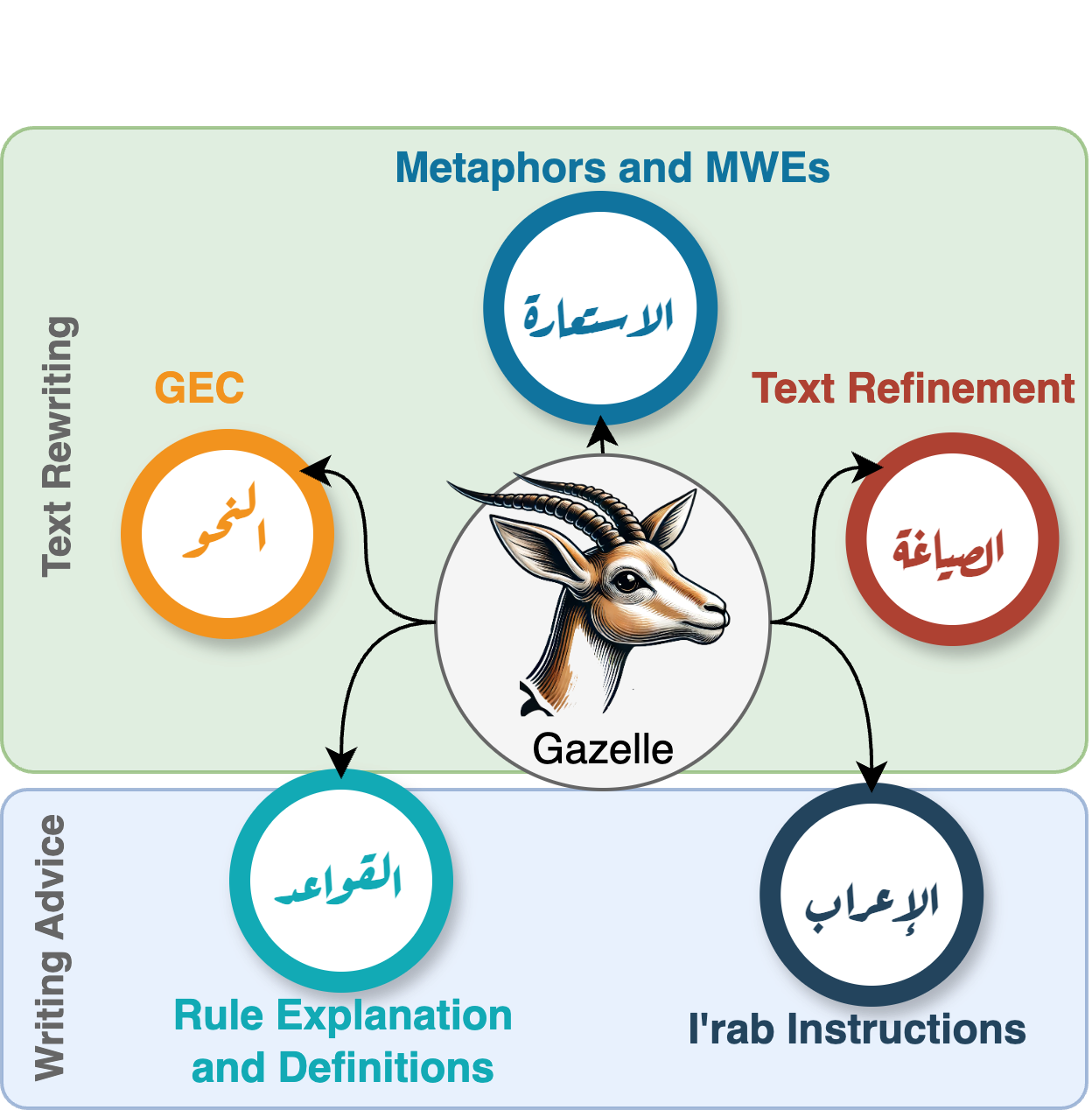}
  \caption{Task categorization for \textit{Gazelle}. \textbf{GEC}: Grammatical Error Correction \AR{\small{(التصحيح النحوي)}}; \textbf{MWEs}: Multi-Word Expressions.}
  \label{fig:trj_overview} 
\end{figure}
In recent years, the landscape of writing assistance has been transformed by rapid advances in generative AI, particularly through the development of LLMs~\cite{ippolito2022creative, lee2024design}. These neural models, trained on the intricacies of language by processing vast amounts of text, have led to the creation of AI-powered tools designed to provide comprehensive writing assistance across a wide range of tasks~\cite{zhang2022opt,chowdhery2023palm}. Models like GPT-$3$ and GPT-$4$~\cite{brown2020language,achiam2023gpt} have demonstrated impressive capabilities in both understanding and generating text~\cite{bubeck2023sparks}. These LLMs have significantly expanded the potential for "Human-AI co-writing," where AI systems provide flexible support that extends far beyond simple grammar and spell-checking to polishing texts and generating completely new content, significantly enhancing writing assistance for diverse writing tasks~\cite{li2024value}.

 \begin{figure*}[t]
   \centering
   \includegraphics[width=\linewidth]{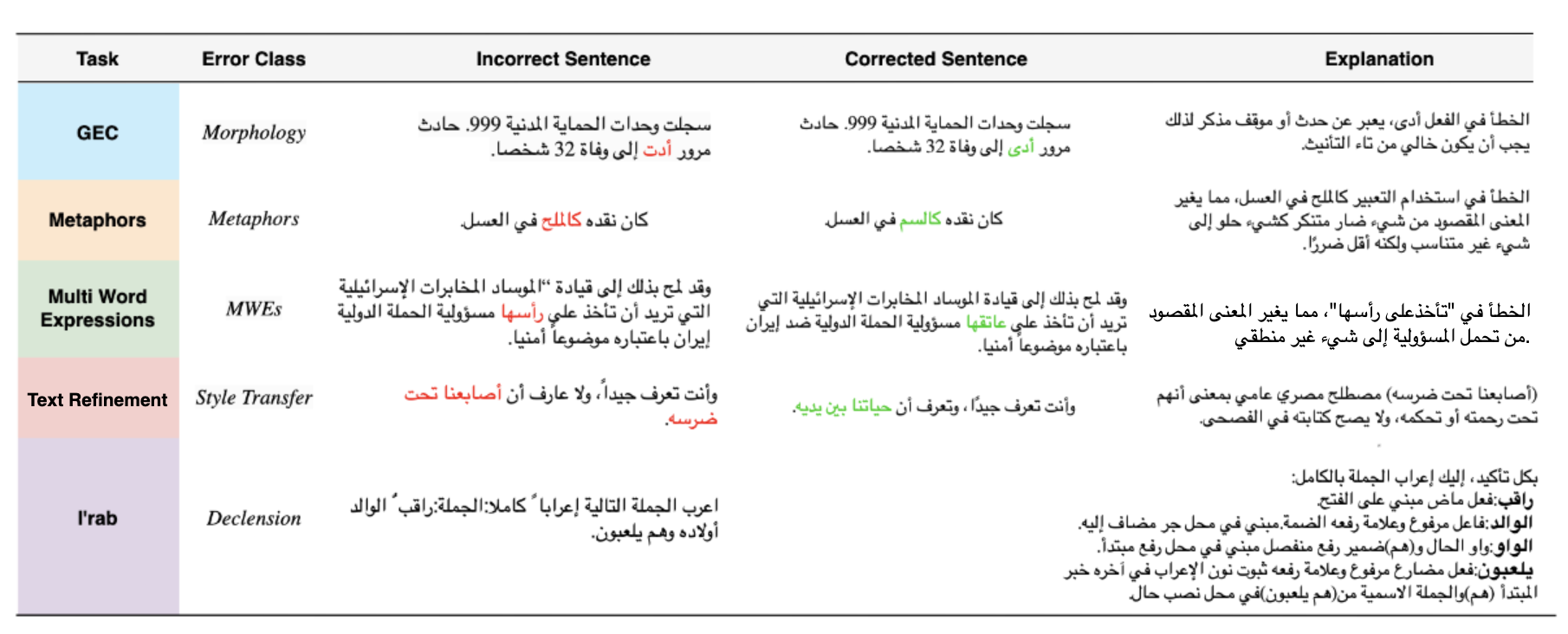}
   \caption{Examples of corrections and explanations for various Arabic writing tasks included in \textit{Gazelle}. For instructions in English, see Table~\ref{fig:subtasks}}
   \label{fig:figure_1} 
 \end{figure*}

However, underrepresented languages like Arabic often face substantial challenges in developing advanced AI tools, including writing assistance, largely due to the scarcity of relevant training data. 
In this work, we seek to bridge this gap by introducing \textit{Gazelle}, a dataset specifically curated for Arabic writing, designed to advance the development of AI-powered writing tools for underrepresented languages. \textit{Gazelle} is a manually curated instruction-style dataset crafted in both English and Arabic and organized around two primary themes. The first theme, {\bf text rewriting}, aims at supporting learners in rewriting their texts across different linguistic levels. This encompasses  \textit{(i) Grammatical Error Correction} (GEC) at various levels such as orthographic, morphological, syntactic, and semantic with catgeories beyond those proposed in the Arabic Learner Corpus (ALC)~\cite{alfaifi2014evaluation} adopted in previous works. Text rewriting also involves corrections of deeper semantic levels beyond traditional GEC that cover  \textit{(ii) Metaphors and Multi-word Expressions} (MWEs) and \textit{(iii) Text Refinement}, which focuses on improving texts by replacing dialectal sequences with their equivalent standard forms and fixing poor language. 

The second theme, {\bf writing advice}, covers \textit{(iv) Rule Explanation and Definitions} as well as \textit{(v) I'rab}. In Arabic, I'rab or declension refers to the variation in a word's form that indicates its grammatical case and role within a sentence. In \textit{Gazelle}, the I'rab category guides learners through the different grammatical roles within a text, such as identifying when a word takes on a nominal case due to its role as a subject in a sentence. Across its various categories, \textit{Gazelle} also provides explanations for the suggested corrections. Figure~\ref{fig:trj_overview} outlines the main tasks covered in \textit{Gazelle}, and Figure~\ref{fig:figure_1} illustrates each of these tasks.

Our contributions can be summarized as follows: \textbf{(1)} We introduce a comprehensive, manually curated dataset specifically for Arabic, covering two different writing themes across five distinct tasks. 
\textbf{(2)} We expand the Arabic Learner Corpus (ALC)~\cite{alfaifi2014evaluation} error taxonomy by developing a fine-grained classification, including adding sub-classes and detailed sub-subclasses under the main error categories. Furthermore, we generate synthetic data from the Arabic Tree Bank (ATB)~\cite{maamouri2004penn} that aligns with our new classification and sub-classes for specific error types in the ALC error taxonomy. \textbf{(3)} We conduct a comparative evaluation of current leading LLMs —\texttt{GPT-$4$}~\cite{brown2020language}, \texttt{GPT-$4o$}~\cite{achiam2023gpt}, \texttt{Cohere Command R+}~\footnote{https://docs.cohere.com/docs/command-r-plus}, and \texttt{Gemini $1.5$ Pro}~\cite{reid2024gemini}—focusing on their capabilities in different Arabic writing tasks.

The rest of this paper is organized as follows: In Section~\ref{RW_updated}, we provide an overview of related work. Section~\ref{sec:writing_assistance_instructions} introduces our manually curated datasets for Arabic writing assistance. In Section~\ref{sec:comparison_with_QALB}, we provide a comparison of Gazelle with other GEC datasets. Section~\ref{sec:eval} presents our evaluation of the frontier LLMs on Arabic writing tasks. We also detail our human evaluation methodology and results. We conclude in Section~\ref{sec:conclusion}.

\section{Related Work}\label{RW_updated}

\noindent\paragraph{LLMs for Writing Assistance.}
The advent of LLMs and their enhancement through data augmentation and instruction tuning have significantly advanced automated writing assistance tools~\cite{wei2022finetuned,ouyang2022training,raheja2023coedit}. Models trained on extensive datasets of human-written instructions, using methods like instruction fine-tuning, have proven to enhance their ability to generalize across various tasks by leveraging large, diverse sets of instructional data~\cite{zhang2023instruction}. Models like FLAN-T5~\cite{chung2024scaling} and GPT-$4$~\cite{bubeck2023sparks,peng2023instruction,chung2024scaling} demonstrate the progress and capabilities of incorporating a wide array of tasks during fine-tuning to achieve robust performance and adaptability in various writing applications~\cite{sanh2021multitask,liu2022few}.
We provide detailed examples of recent advancements in instruction-tuned models and their application on writing assistance in Appendix~\ref{apndx:related_works}.

\noindent\paragraph{Progress in Arabic Writing Tasks.}
Arabic writing tasks are complex due to its linguistic and morphological diversity, compounded by orthographic ambiguities from optional diacritics and dialectal variations~\cite{abdul-mageed-etal-2020-nadi, belkebir2021automatic}. Despite these challenges, significant progress has been made in terms of model development and dataset developments across different writing tasks such as GEC~\cite{zaghouani-etal-2014-large, habash-palfreyman-2022-zaebuc,kwon2023beyond}, text simplification~\cite{hazim-etal-2022-arabic,alhafni2024samer} and text summarization~\cite{Al-Maleh_summarization, Lagrini2021ANA}. Further details about the progress in Arabic writing tasks are provided in Appendix~\ref{apndx:related_works}.

\noindent\paragraph{Explainable Writing Systems.}
Recent advancements in explainable writing systems have greatly improved the transparency and alignment of tools with instructional goals across various tasks. In GEC, tools like ERRANT~\cite{bryant-etal-2017-automatic} provide clear error annotations and edit boundaries, enhancing users' understanding of grammatical mistakes. In text simplification and feedback generation, systems strive to offer precise and relevant explanations, ensuring that users receive actionable and comprehensible feedback~\cite{agarwal2022explain}. These efforts highlight the importance of creating tools that deliver understandable insights and ensure alignment with instructional goals \cite{kunz2024properties}. We provide further details on the progress of explainable writing systems in Appendix~\ref{apndx:related_works}. However, to the best of our knowledge, no prior work has investigated instruction-tuned LLMs' ability to provide explanations and definitions for diverse writing tasks, organized around distinct pedagogical themes.

\section{Gazelle}
\label{sec:writing_assistance_instructions}
As mentioned in Section~\ref{sec:intro}, {\it Gazelle} comprises two main themes, text rewriting and writing advice, across \textit{five} distinct types of writing assistance instructions that we manually curate. {\it Gazelle} encompasses parallel data with instructions crafted in both Arabic and English, covering both input and output, allowing users to fine-tune models for these various tasks bilingually. Except for the I'rab and Arabic grammatical rules and definitions datasets, we manually translate the Arabic explanations and instructions into English. 
To ensure thorough coverage, we curate instructions for each of these tasks separately from several publicly available online sources~\footnote{See Appendix~\ref{online_resources} for example sources we use.}. Table~\ref{data_stats_gec} presents the statistics for text rewriting tasks, while Table~\ref{data_stats_other} provides statistics for tasks related to writing advice. Examples of these tasks are illustrated in Figure~\ref{fig:figure_1}.
For the English translations of the GEC explanations and definitions task dataset in Table~\ref{data_stats_gec}, we leverage GPT-$4o$~\footnote{https://openai.com/index/hello-gpt-4o/} to accurately translate the manually curated Arabic content. This process involved iteratively prompting GPT-$4o$ and refining the generated 
through multiple rounds of manual review to ensure the translations were both accurate and contextually appropriate, allowing us to produce high-quality English translations.  We explain each of our instruction categories next.
\begin{table}[t]
\centering
\resizebox{\columnwidth}{!}{%
\begin{tabular}{lcccc}
\toprule
\textbf{Task} & \textbf{Category} & \textbf{Source} & \textbf{Language} & \textbf{Total} \\
\midrule
\multirow{8}{*}{\textbf{GEC}} & Orthographic & Online Sources & \multirow{8}{*}{\centering Arabic} & 637 \\
 & Syntax & Online Sources &  & 377 \\
 & Semantic & Online Sources &  & 362 \\
 & Morphology & Online Sources &  & 55 \\
 & Punctuation & Online Sources &  & 28 \\
 & Split & ATB &  & 100 \\
 & Merge & ATB &  & 100 \\
\midrule
\textbf{GEC\_ATB} &  & ATB & Arabic & 1,433 \\
\midrule
\textbf{GEC + Definitions} &  & Online Sources & Arabic & 290 \\
\midrule
\textbf{GEC + Explanations} &  & Online Sources & Arabic & 1,659 \\
\midrule
\textbf{MWE} &  & PARSEME dataset & Arabic & \multirow{2}{*}{\centering 98} \\
 &  &  & English & \\
\midrule
\textbf{Text Refinement} &  & In-house Project & Arabic & \multirow{2}{*}{\centering 770} \\
 &  &  & English & \\
\bottomrule
\end{tabular}%
}
\caption{Data statistics for \textit{text rewriting} tasks.}
\label{data_stats_gec}
\end{table}

\begin{table}[t]
\centering
\resizebox{\columnwidth}{!}{%
\begin{tabular}{lcccc}
\toprule
\textbf{Task} & \textbf{Source} & \textbf{Language} & \textbf{Total} \\
\midrule
\multirow{2}{*}{\textbf{Rule Explanations \& Definitions}} & Online Sources & Arabic & 1,445 \\
 & Online Sources & English & 1,005 \\
\midrule
\textbf{I'rab} & Online Sources & Arabic & 1,090 \\
\bottomrule
\end{tabular}%
}
\caption{Data statistics for \textit{writing advice} tasks.}
\label{data_stats_other}
\end{table}



\subsection{Text Rewriting}
\subsubsection{GEC}
\noindent\paragraph{Fine-Grained Error Taxonomy.}
The original ALC error taxonomy comprises 29 error tags categorized under five main error classes: \textbf{\textit{Orthographic}}, \textbf{\textit{Morphology}}, \textbf{\textit{Syntax}}, and \textbf{\textit{Semantics}}. Additionally, \citet{Belkebir2021AutomaticET} introduced two more error classes: \textbf{\textit{Split}} and \textbf{\textit{Merge}}. Building upon this foundation, we expand the original ALC error taxonomy to develop a fine-grained grammatical error taxonomy. 

Our proposed taxonomy introduces two more sub-classes to the original ALC error categories: \textit{Closed Class Error} under \textbf{\textit{Syntax}} and \textit{Special Expression Error} under \textbf{\textit{Semantic}}, bringing the total to 31 error tags. \textit{Closed class} errors address issues with closed class items, such as \textit{the five nouns}, \textit{the five verbs}, and \textit{pronouns}. \textit{Special expression errors} covers the use of collocations, multi-word expressions, and other commonly misused terms 
Additionally, each of our new error categories include several sub-subclasses. For example, the sub-classs \textit{Number} under \textbf{\textit{Syntax}} which pertains to the correct use of singular, dual, and plural forms in nouns, verbs, and pronouns is further divided into several sub-subclasses. Our newly-developed categorization scheme incorporates extensive knowledge of Arabic grammar, aiming to provide users with not only the category of grammatical error but also its sub- and sub-subclasses. 
Table~\ref{extended_acl_errors_appdx} presents a detailed overview of all the newly introduced categories. It comprises 7 classes, each with an average of 9 sub-classes~\footnote{We exclude all error tags within the \textit{Other} error class (e.g., OO, XO, SO, MO) to avoid complications.}.  

\noindent\paragraph{GEC Instructions.}
We develop a set of 3,382 bilingual instructions for GEC, based on our newly extended error taxonomy. This comprehensive taxonomy includes a wide range of error sub-classes and sub-subclasses, aiding identification of specific errors. We manually curate these sentences from various Arabic learning online sources to reflect the detailed error categories in our extended taxonomy. Each sentence pair includes grammatical errors along with their corrected versions and explanations, as shown in Table~\ref{tab:Extended Error Taxonomy}. Examples of these detailed error sub-classes and their coverage are provided in Appendix~\ref{writing_assistance_instructions}.

\noindent\paragraph{GEC Error Explanations.}
We introduce a dataset containing detailed explanations for each type of error sub-class and sub-subclass, comprising a total of 1,659 entries sourced from websites and books. This dataset enables users not only identify the type of error but also understand the explanations behind it. The explanations are provided in both Arabic and English. We provide examples of these detailed error explanations in Appendix~\ref{writing_assistance_instructions}.

\noindent\paragraph{ATB Synthetic Data.}
ATB is a collection of texts annotated for morphosyntax with English glosses. We utilize all parts of the ATB to generate 100 examples with a total of 1,633 covering the \textbf{\textit{Orthographic}}, \textbf{\textit{Syntax}}, \textbf{\textit{Split}}, \textbf{\textit{Merge}}, \textbf{\textit{Morphological}}, and \textbf{\textit{Punctuation}} error classes along with their sub-classes. More details on the examples generated are provided in Appendix~\ref{writing_assistance_instructions}.

Generating synthetic data from ATB allows us to efficiently acquire real-world data with human labels. We then use templates to format these examples into an instruction format.

For each error type, we follow specific rules that dictate how errors are created, ensuring consistency throughout the process. We prioritize aligning our error generation with those made by humans, rather than relying on random or arbitrary variations. For instance, with OM (Orthographic Missing) errors, where characters are missing, we simulate typical human mistakes by deliberately omitting characters that are commonly omitted in real-world usage. Appendix \ref{atb_synthetic_data_appdx} provides detailed information on the methodology we follow to generate these synthetic errors from ATB.

\subsubsection{Metaphors and MWEs}
MWEs are diverse and arbitrary word combinations that co-occur frequently. They include idiomatic phrases, compound words, and other constructions with varying levels of transparency and fixedness, all sharing the characteristic of crossing word boundaries~\cite{sag2002multiword}. Our dataset comprises 330 sentences featuring incorrect usage of metaphors or MWEs, along with corrected versions and explanations of the errors. The sentences with metaphors were manually collected from online resources. For MWE sentences, we randomly selected 98 sentences from the PARSEME dataset~\cite{mohamed2022annotating} a collection of verbal multiword expressions in Arabic. To generate sentences with errors, we use GPT-$4o$ to modify correct metaphors and MWEs into incorrect ones. We provide the correct sentence and instruct it to alter the metaphor or MWEs. 

\subsubsection{Text Refinement}
Text Refinement refers to improving the quality of MSA by addressing informal or colloquial words in a text or by enhancing texts written in poor language. We create a set of instructions comprising a total of 770 sentences (we collect these sentences from an in-house project) focused on detecting informal words and improving text quality, along with explanations of these errors. Examples regarding this task can be found in Appendix~\ref{writing_assistance_instructions}.

\subsection{Writing Advice}
\subsubsection{Rule Explanation and Definitions}
We develop a bilingual dataset consisting of $2,450$ examples that provide explanations and definitions of grammatical rules in both Arabic and English. This dataset covers a wide range of topics in Arabic grammar, from basic concepts such as nouns, verbs, negation, and sentence structures to advanced topics like verb inflections, formations, and stylistic nuances. Each entry offers comprehensive definitions and illustrative examples to enhance understanding. Contextual examples are also included to demonstrate the practical application of these rules. Illustrative examples can be found in Figure~\ref{fig:subtasks}.

\subsubsection{I'rab}
I'rab, or declension, is a term in Arabic that refers to the system of nominal, adjectival, or verbal suffixes used to mark the grammatical case of a word. These suffixes helps identify the function of a word within a sentence~\cite{hamani2014phenomenon}.
We design 1,090 instructions that provide the full I'rab of sentences, covering a wide range of topics in Arabic grammar, as detailed in Figure \ref{fig:subtasks}.

\section{Gazelle in Comparison} 

\label{sec:comparison_with_QALB}
The QALB dataset comprises manually corrected Arabic texts, including MSA commentaries from Aljazeera by native speakers (L1) and texts by non-native learners (L2)\cite{rozovskaya2014columbia, Mostefa2015TECHLIMEDQALBSharedT2}. The ZAEBUC corpus, on the other hand, contains essays written by first-year students at Zayed University in the UAE\cite{habash-palfreyman-2022-zaebuc}. While both the QALB and ZAEBUC corpora include multiple errors per sentence or paragraph, facilitating automatic evaluation, our datasets are curated with a different focus.

We manually curate the \textit{Gazelle} dataset to target specific error types within the ALC error taxonomy by extending the existing error categories and incorporating our own extended categories. Each sentence, sourced from various Arabic learning online sources, represents a specific subclass or sub-subclass of our extended error taxonomy and contains only one type of error along with explanation of this type of error. This targeted approach makes our dataset particularly useful for learners aiming to understand and correct specific types of errors. We provide a comparison of \textit{Gazelle} dataset to the QALB and ZAEBUC corpora in Appendix~\ref{comparsion}

\section{Evaluating Leading LLMs on Gazelle}
\label{sec:eval}
We evaluate \textit{Gazelle} on current leading LLMs, namely GPT-$4$, GPT-$4o$, Gemini $1.5$ Pro, and Cohere Command R+, to assess their capabilities across our proposed tasks in Arabic. We test the models under a zero-shot setting~\cite{sanh2021multitask}, allowing us to assess the models' inherent capabilities in handling our proposed writing tasks. For each task, we initially tested the prompts and iteratively refined a zero-shot prompt tailored to each subtasks. Further details on our prompt design process, along with examples used for evaluation, are provided in Appendix~\ref{apdx:prompt_design}.

\subsection{Human Evaluation}
We conduct human evaluation to assess the performance of the aforementioned LLMs on \textit{Gazelle's} five subtasks. Developing evaluation criteria for these subtasks in Arabic writing was challenging due to the lack of existing standard measures. Consequently, we create a list of suitable measures for all subtasks and iteratively refine them. Each criterion was clearly defined to ensure both comprehensiveness and clarity.

For evaluation, we ask four native Arabic speakers to assess the models' performance by randomly sampling a subset of 20 sentences from each task, resulting in a total of 100 sentences. Annotators were then asked to evaluate the models' outputs for each task using the metrics we developed, which include \textit{accuracy}, \textit{clarity}, \textit{helpfulness}, \textit{appropriateness}, \textit{correctness}, \textit{consistency}, \textit{depth} of explanation, \textit{quality} of text refinement, \textit{fluency}, and \textit{relevance}. Each response was scored on a scale from 1 to 5, reflecting on how well it met the evaluation criteria. Detailed descriptions of the evaluation scales are provided in Appendix~\ref{apdx:matrices}.

\begin{figure}
    \centering
    \includegraphics[width=0.8\linewidth]{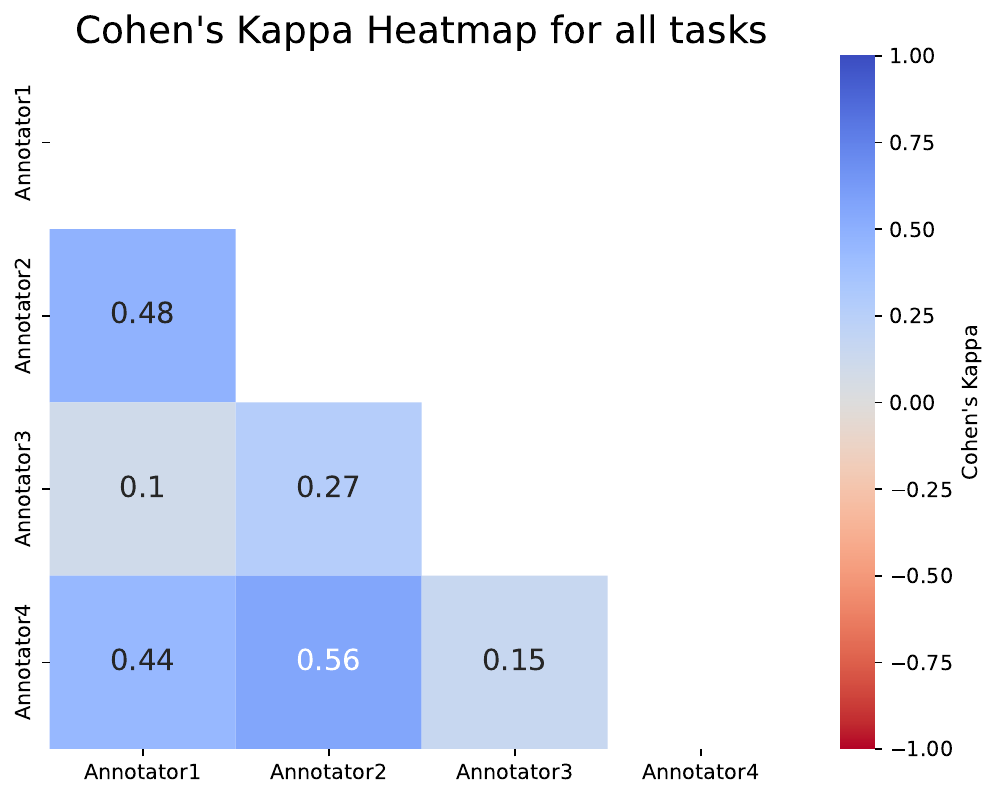}
    \caption{Overall inter-annotator agreement for human evaluation measured by Cohen's Kappa.}
    \label{fig:overall-agreement}
\end{figure}

We calculate the Cohen’s Kappa scores for inter-annotator agreement among four annotators across all tasks. Figure \ref{fig:overall-agreement} displays the overall inter-annotator agreement, which varies across annotator pairs, with some pairs showing moderate agreement and others showing slight to fair agreement. Individual inter-annotator agreements are provided in Appendix \ref{fig:agreement}.

\subsection{Results}
\begin{figure*}[t]
    \centering
    \begin{subfigure}[b]{0.45\textwidth}
        \centering
        \includegraphics[width=\textwidth]{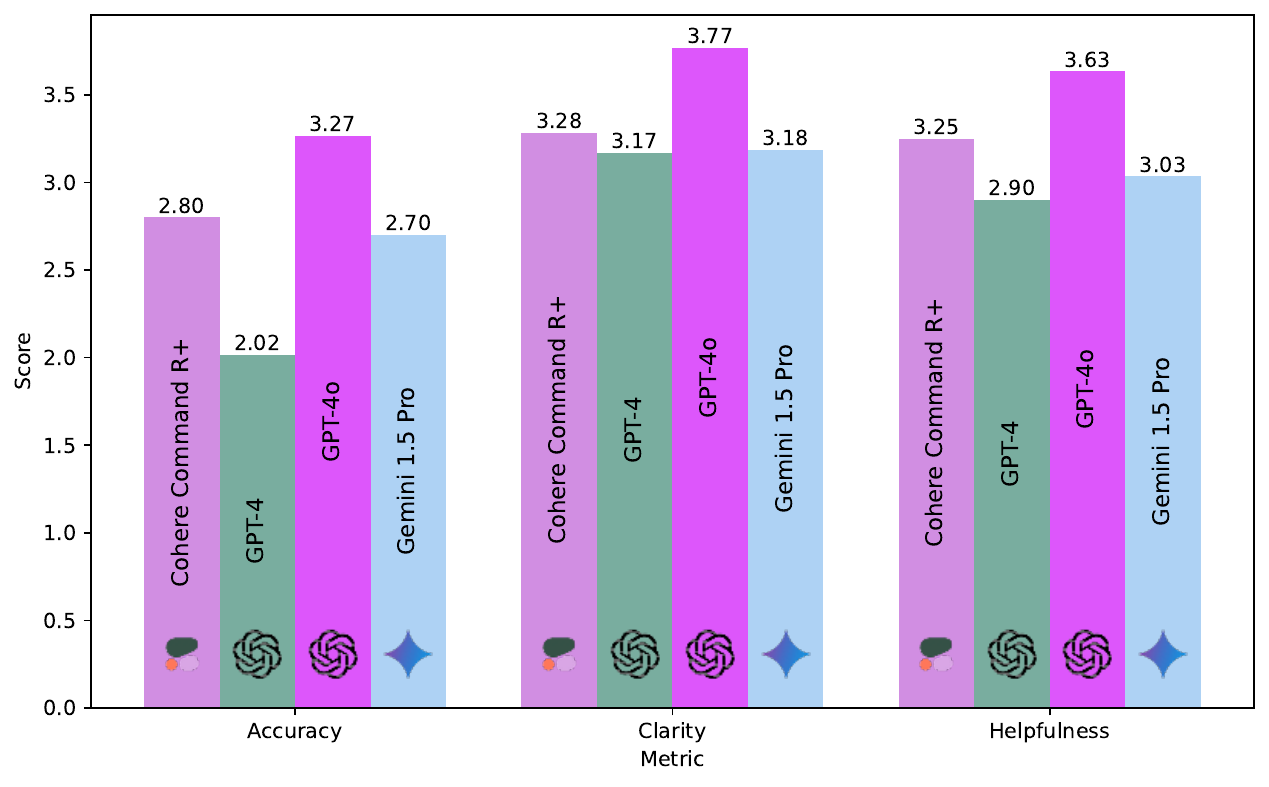}
        \caption{GEC + Explanation}
        \label{fig:gec_result}
    \end{subfigure}
    \hfill
    \begin{subfigure}[b]{0.45\textwidth}
        \centering
        \includegraphics[width=\textwidth]{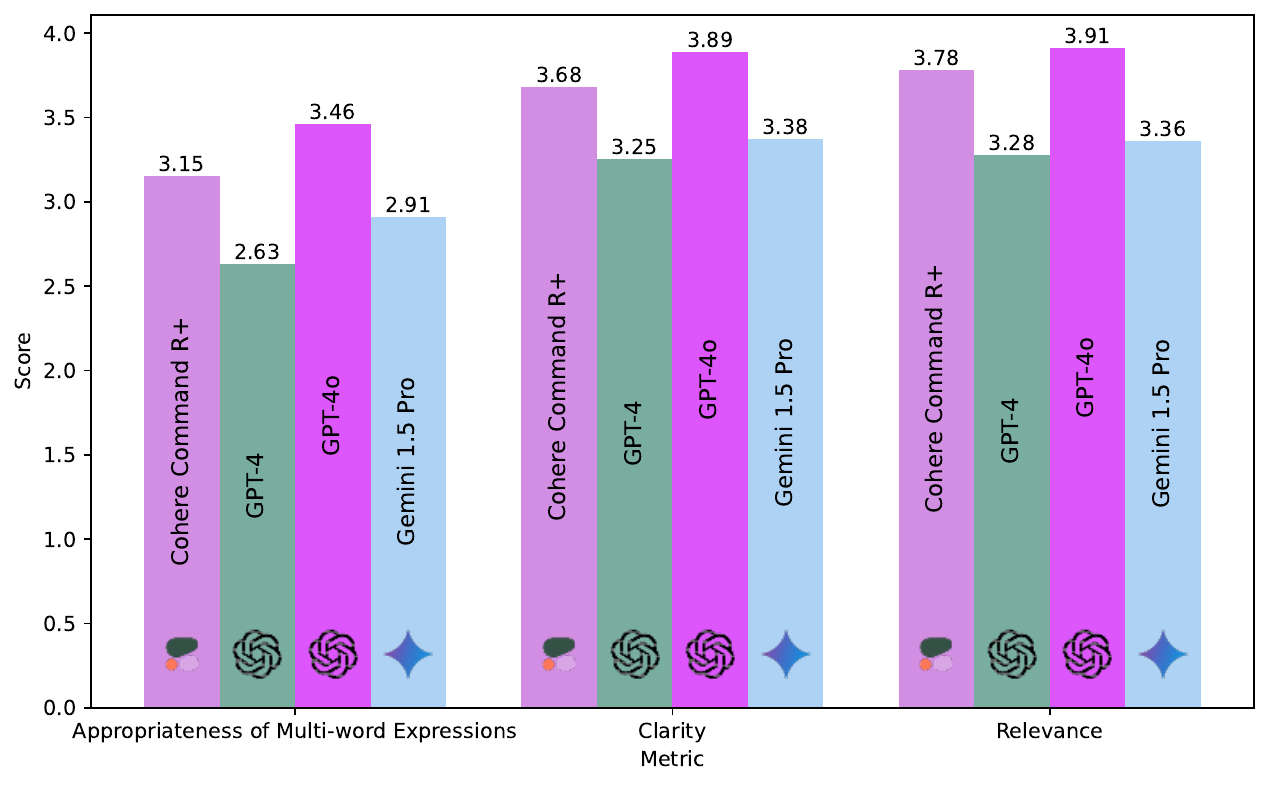}
        \caption{MWEs}
        \label{fig:mwe_result}
    \end{subfigure}
    \vfill
      \begin{subfigure}[b]{0.45\textwidth}
        \centering
        \includegraphics[width=\textwidth]{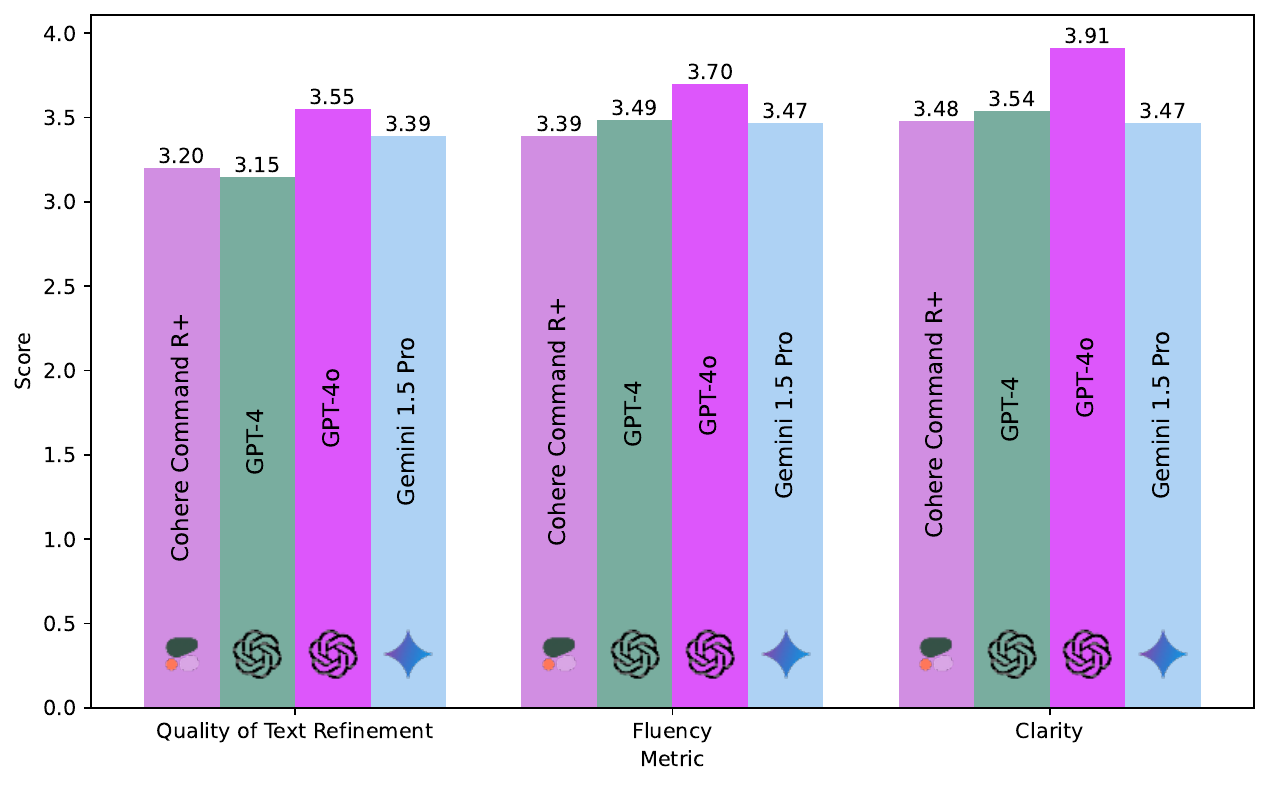}
        \caption{Text Refinement}
        \label{fig:text_result}
    \end{subfigure}
    \begin{subfigure}[b]{0.45\textwidth}
        \centering
        \includegraphics[width=\textwidth]{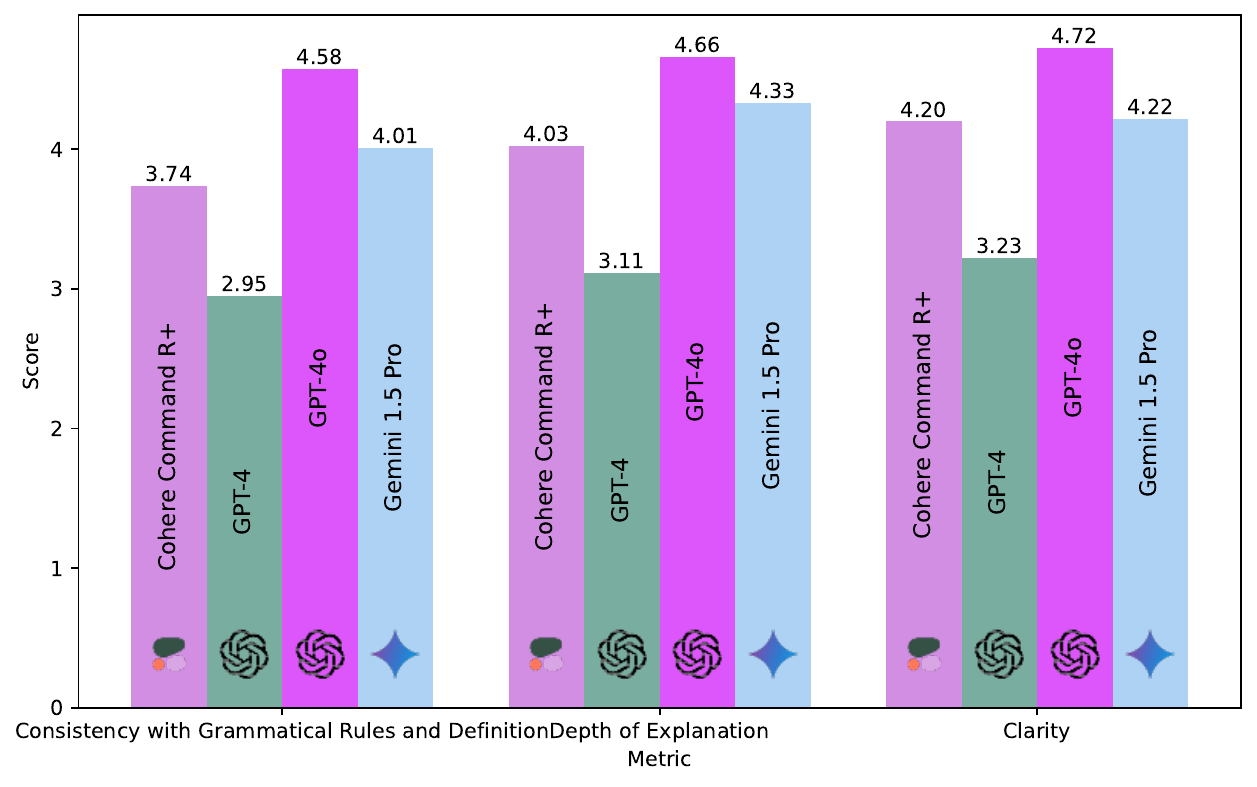}
        \caption{Grammatical Rules and Definitions}
        \label{fig:rules_result}
    \end{subfigure}
    \hfill
   
    \begin{subfigure}[b]{0.45\textwidth}
        \centering
        \includegraphics[width=\textwidth]{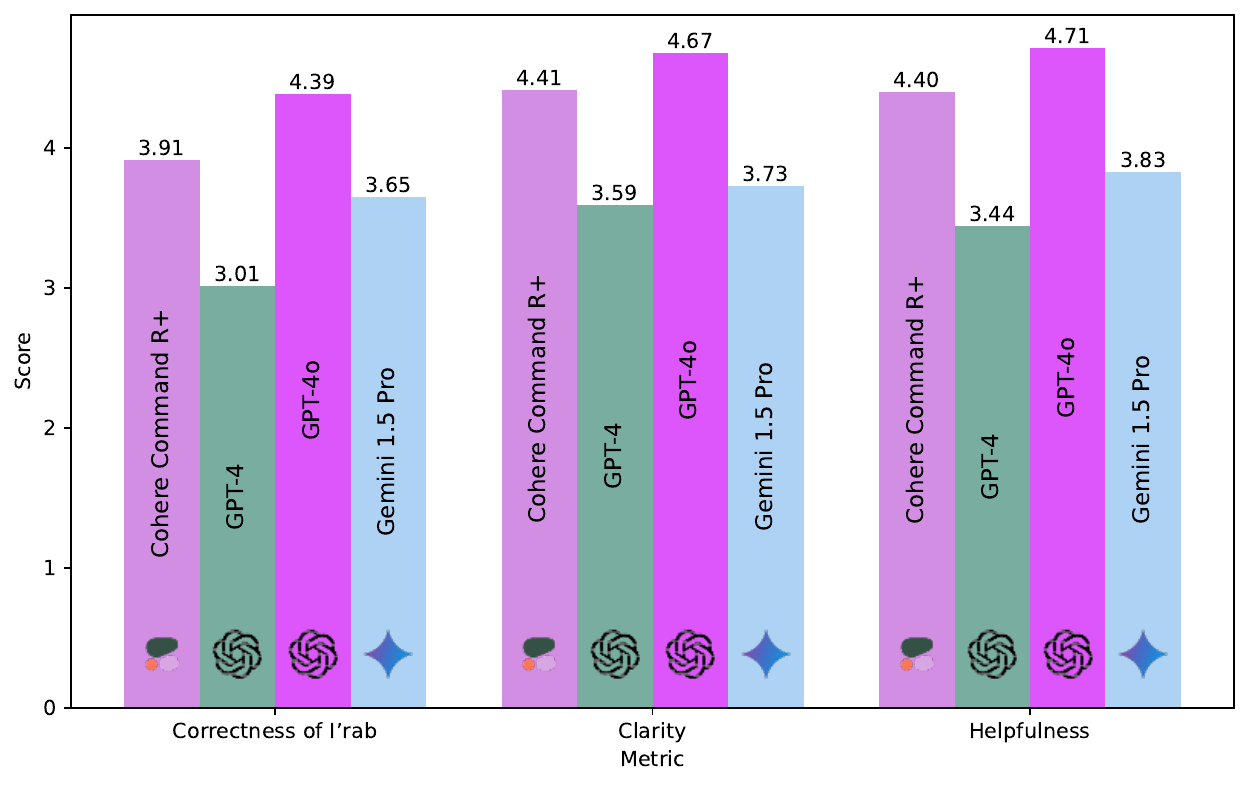}
        \caption{I'rab}
        \label{fig:iraab_result}
    \end{subfigure}
    \hspace{0.05\textwidth} 
    \caption{Results of human evaluation for four LLM models: GPT-$4$, GPT-$4o$, Cohere Command R+, and Gemini $1.5$ Pro on five subtasks in \textit{Gazelle}.} 
    \label{fig:eval_results}
\end{figure*}

We present the results of our evaluation in Figure~\ref{fig:eval_results}. The performance of each model on the outlined writing tasks is discussed below.

\noindent\paragraph{GEC and Explanation Tasks.} For the GEC and explanation tasks, we evaluate the performance of the models based on three criteria: \textit{accuracy}, \textit{clarity}, and \textit{helpfulness}.
Figure~\ref{fig:gec_result} displays the evaluation result for this task. 

Overall, \texttt{GPT-$4o$} demonstrates superior performance in the GEC task, excelling in all criteria. \texttt{Cohere Command R+} and \texttt{Gemini $1.5$ Pro} provide balanced results, whereas \texttt{GPT-$4$}, while maintaining a reasonable level of \textit{clarity}, falls short in \textit{accuracy} and \textit{helpfulness}. \\

\noindent\paragraph{MWEs.}
In the MWEs task, we evaluate the four models based on \textit{clarity}, \textit{appropriateness} of multi-word expressions, and \textit{relevance}. Figure~\ref{fig:mwe_result} shows result of the evaluation for this task.

Overall, \texttt{GPT-$4o$} outperforms all models in the MWEs task, delivering the highest scores in all criteria. Both \texttt{Cohere Command R+} and \texttt{Gemini $1.5$ Pro} perform well, particularly in \textit{clarity} and \textit{relevance}, while GPT-$4$ lags in appropriateness but maintains reasonable levels of \textit{clarity} and \textit{relevance}. \\

\noindent\paragraph{Text Refinement.}
We evaluate the text refinement task based on \textit{clarity}, \textit{quality} of text refinement, and \textit{fluency}. Figure~\ref{fig:text_result} presents result of the evaluation for this task.

\texttt{GPT-$4o$} excels in the Text Refinement task, leading with superior scores in \textit{clarity}, \textit{quality} of text refinement, and \textit{fluency}. Both \texttt{Cohere Command R+} and \texttt{Gemini $1.5$ Pro} show balanced and consistent performance, while \texttt{GPT-$4$} also performs well, particularly in \textit{fluency}. \\

\noindent\paragraph{Grammatical Rules and Definitions.}
For this task, we evaluate the models based on three criteria: \textit{clarity}, \textit{consistency} with grammatical rules and definitions, and \textit{depth} of explanation. Figure~\ref{fig:rules_result} illustrates the result for this task.

In summary, \texttt{GPT-$4o$} performs exceptionally in the grammatical rules and definitions task, scoring highest in all criteria. \texttt{Gemini $1.5$ Pro} follows closely, offering strong and balanced performance. \texttt{Cohere Command R+} also performs well, particularly in \textit{clarity} and \textit{depth}, while \texttt{GPT-$4$} shows the lowest scores across all criteria. \\

\noindent\paragraph{I'rab.}
We evaluate the models on three criteria for the I'rab task: \textit{clarity}, \textit{helpfulness}, and \textit{correctness} of I’rab. Figure~\ref{fig:iraab_result} presents result of the evaluation for this task.

Overall, \texttt{GPT-$4o$} dominates the I’rab task with superior scores in \textit{clarity}, \textit{helpfulness}, and \textit{correctness}. \texttt{Cohere Command R+} also performs well, especially in \textit{clarity} and \textit{helpfulness}. \texttt{Gemini $1.5$ Pro} shows balanced results, while \texttt{GPT-4} has the lowest scores across all criteria. \\

\subsection{Discussion}

\noindent\paragraph{Cohere Command R+.} \texttt{Cohere Command R+} excels in delivering clear and helpful explanations, particularly in handling MWEs and I’rab writing tasks. It consistently offers understandable and relevant responses in these areas. However, its performance in GEC tasks is less robust, with lower \textit{accuracy} and \textit{correctness} compared to other models. We provide detailed statistical evaluation metrics for the five subtasks across all models in~Table \ref{tab:stats evaluation}

\noindent\paragraph{GPT-4.} \texttt{GPT-4} demonstrates the weakest overall performance among all the models evaluated. While it does provide reasonable \textit{clarity} and \textit{relevance} in its responses, it struggles with the \textit{accuracy} and \textit{depth} required for tasks such as GEC, explaining grammatical rules, and providing GEC explanation. It scores the lowest compared to the other models in these critrea. Although \texttt{GPT-4}’s performance is stable, it does not excel in any specific task.

\noindent\paragraph{GPT-4o.} \texttt{GPT-4o} stands out as the top-performing model across all tasks. It consistently achieves the highest scores by demonstrating superior \textit{clarity}, \textit{correctness}, and \textit{depth} in its responses. \texttt{GPT-4o} reliably provides the most \textit{accurate}, \textit{relevant}, and \textit{helpful} outputs, making it the most effective model for comprehensive Arabic writing assistance. Its performance is robust and reliable, with minimal weaknesses observed.

\noindent\paragraph{Gemini 1.5 Pro.} \texttt{Gemini 1.5 Pro} delivers a balanced and strong performance across all tasks. It particularly excels in explaining grammatical rules and providing detailed and consistent responses. Although it performs well overall, it does not quite reach the high performance levels of \texttt{GPT-4o}. In tasks involving MWEs and I’rab, \texttt{Gemini 1.5 Pro} shows room for improvement, where it is less effective compared to \texttt{GPT-4o}. \\

\subsection{Human Analysis of the Models}
We outline insights from our human analysis, offering a detailed evaluation of how each model handles specific linguistic challenges across all tasks.

\noindent\paragraph{GEC and  explanation task.}
Models frequently corrected sentences but occasionally struggled with accurate explanations or identifying all errors, highlighting their limitations in grasping Arabic grammar complexities. The task of applying grammatical rules and definitions revealed varying proficiency among the models, showcasing their differing capabilities in handling Arabic grammar.

\noindent\paragraph{MWEs and Metaphors.}
All models struggled with handling MWEs and metaphors. This difficulty was evident in models' ability to identify errors and provide accurate metaphors or MWEs, highlighting their inability to fully capture the linguistic richness of Arabic.

The complexity of handling MWEs and metaphors in Arabic stems from their nuanced and context-dependent nature. Unlike single-word expressions, MWEs and metaphors often convey meanings that cannot be directly inferred from the individual words. Furthermore, the cultural and contextual connotations embedded in Arabic expressions add another layer of difficulty for the models. 

\noindent\paragraph{Text Refinement.}
In the text up-refinement task, a creative exercise, models showed diverse capabilities. Some excelled in enhancing stylistic elements such as \texttt{GPT-$4o$} and \texttt{Gemini 1.5 Pro}, while others focused on basic corrections as in the case of \texttt{GPT-4}. 

\noindent\paragraph{I'rab.}
The I’rab task, requiring deep linguistic knowledge, revealed mixed abilities in applying these rules. Models like \texttt{Cohere Command R+} and \texttt{GPT-$4o$} performed well, while others showed limitations in providing a fully correct I'rab of the sentences.

Overall, this analysis highlights the variability in model performance, showcasing each model's strengths in specific areas and their limitations in understanding and applying Arabic grammar and stylistic enhancements.

\section{Conclusion}
\label{sec:conclusion}
In this work, we present \textit{Gazelle }a comprehensive dataset and evaluation framework to enhance Arabic writing assistance tools. Our manually fine-grained dataset incorporates five writing subtasks under two main themes.  Our analysis uncovers both strengths and limitations of the most leading LLMs in tackling the unique challenges of Arabic writing. These insights highlight the need for continuous model training and dataset enrichment to address the complexities of Arabic language processing.

Our study showcases the varied capabilities of models in handling Arabic grammar and stylistic nuances, providing a foundational step towards more advanced 'Human-AI co-writing' tools for Arabic writing. Future research may include on expanding the dataset and integrating diverse linguistic features to develop robust and reliable models for mastering Arabic writing.

Overall, our work serves as a stepping stone towards creating more effective Arabic writing assistance tools and advancing the field significantly.
\section*{Limitations}
We identify the following limitations in this work:
\begin{itemize}

    \item  The dataset may not encompass the full diversity of the Arabic language, including its many dialects and regional variations, limiting the generalizability of the results.
    
    \item  The study focuses on zero-shot settings, which highlight models' inherent capabilities but not their performance with fine-tuning. Future work may include fine-tuning for enhanced performance.
    \item The human evaluation process is subjective and may be influenced by individual biases, particularly in tasks like text up-scaling and error explanation.
    \item  The study evaluates four specific LLMs. As models evolve, re-evaluation will be necessary to maintain relevance in assessing their performance on Arabic writing tasks.

\end{itemize}

\section*{Ethics Statement}
\noindent\textbf{Encouraging research development and contributing to a collaborative research culture.}
Low-resource languages such as Arabic face significant hurdles in the development of advanced AI tools due to a scarcity of comprehensive datasets. This shortage hampers the ability to train effective models, limiting the creation of sophisticated writing assistance tools for these languages. Our current work aims to bridge this gap by introducing \textit{Gazelle}, a dataset specifically curated for Arabic writing. This dataset is designed to support the development of AI-powered tools that can enhance Arabic language processing. By providing this diverse and detailed dataset, we aspire to stimulate further research and development in Arabic AI tools, fostering a collaborative effort to overcome the challenges faced by low-resource languages.

\noindent\textbf{Advancing Pedagogical Approaches in Arabic Writing.}
With the growing interest in enhancing language proficiency, the accuracy and effectiveness of written communication have become essential for both native speakers and second language learners. LLMs are increasingly utilized as writing assistants, highlighting their significant role in educational tools. Gazelle, supports the development of AI-driven tools that facilitate learning and teaching in Arabic writing. Gazelle focuses on a broad spectrum of writing tasks and aims to assist both non-native Arabic learners and native speakers in understanding the pedagogical aspects of Arabic writing. This approach not only bridges the gap between learners and fluent written communication but also enriches the educational tools available for teaching Arabic. By offering comprehensive resources tailored to diverse writing needs, Gazelle enhances the capability of educational platforms to help users master the intricacies of Arabic writing.

\noindent\textbf{Data privacy.} In relation to the data used in this work, all datasets are publicly available. Therefore, we do not have  privacy concerns.

\section*{Acknowledgments}\label{sec:acknow}
We acknowledge support from Canada Research Chairs (CRC), the Natural Sciences and Engineering Research Council of Canada (NSERC; RGPIN-2018-04267), the Social Sciences and Humanities Research Council of Canada (SSHRC; 895-2020-1004; 895-2021-1008), Canadian Foundation for Innovation (CFI; 37771), Digital Research Alliance of Canada,\footnote{\href{https://alliancecan.ca}{https://alliancecan.ca}} and UBC Advanced Research Computing-Sockeye.\footnote{\href{https://arc.ubc.ca/ubc-arc-sockeye}{https://arc.ubc.ca/ubc-arc-sockeye}}

\bibliography{anthology,custom}

\begin{thebibliography}{58}
\expandafter\ifx\csname natexlab\endcsname\relax\def\natexlab#1{#1}\fi

\bibitem[{Abdul-Mageed et~al.(2020)Abdul-Mageed, Zhang, Bouamor, and Habash}]{abdul-mageed-etal-2020-nadi}
Muhammad Abdul-Mageed, Chiyu Zhang, Houda Bouamor, and Nizar Habash. 2020.
\newblock \href {https://aclanthology.org/2020.wanlp-1.9} {{NADI} 2020: The first nuanced {A}rabic dialect identification shared task}.
\newblock In \emph{Proceedings of the Fifth Arabic Natural Language Processing Workshop}, pages 97--110, Barcelona, Spain (Online). Association for Computational Linguistics.

\bibitem[{Achiam et~al.(2023)Achiam, Adler, Agarwal, Ahmad, Akkaya, Aleman, Almeida, Altenschmidt, Altman, Anadkat et~al.}]{achiam2023gpt}
Josh Achiam, Steven Adler, Sandhini Agarwal, Lama Ahmad, Ilge Akkaya, Florencia~Leoni Aleman, Diogo Almeida, Janko Altenschmidt, Sam Altman, Shyamal Anadkat, et~al. 2023.
\newblock \href {https://cdn.openai.com/papers/gpt-4.pdf} {Gpt-4 technical report}.
\newblock \emph{arXiv preprint arXiv:2303.08774}.

\bibitem[{Agarwal(2022)}]{agarwal2022explain}
Aman Agarwal. 2022.
\newblock \href {{https://arxiv.org/abs/2212.04595}} {Explain to me like i am five--sentence simplification using transformers}.
\newblock \emph{arXiv preprint arXiv:2212.04595}.

\bibitem[{Al~Khalil et~al.(2018)Al~Khalil, Saddiki, Habash, and Alfalasi}]{al-khalil-etal-2018-leveled}
Muhamed Al~Khalil, Hind Saddiki, Nizar Habash, and Latifa Alfalasi. 2018.
\newblock \href {https://aclanthology.org/L18-1366} {A leveled reading corpus of {M}odern {S}tandard {A}rabic}.
\newblock In \emph{Proceedings of the Eleventh International Conference on Language Resources and Evaluation ({LREC} 2018)}, Miyazaki, Japan. European Language Resources Association (ELRA).

\bibitem[{Al-Maleh and Desouki(2020)}]{Al-Maleh_summarization}
Molham Al-Maleh and Said Desouki. 2020.
\newblock \href {https://doi.org/10.1186/s40537-020-00386-7} {Arabic text summarization using deep learning approach}.
\newblock \emph{Journal of Big Data}, 7.

\bibitem[{Alfaifi and Atwell(2014)}]{alfaifi2014evaluation}
Abdullah Alfaifi and ES~Atwell. 2014.
\newblock \href {https://eprints.whiterose.ac.uk/81592/} {An evaluation of the arabic error tagset v2}.
\newblock In \emph{Proceedings of the AACL 2014-The American Association for Corpus Linguistics conference}. The American Association for Corpus Linguistics.

\bibitem[{Alhafni et~al.(2024)Alhafni, Hazim, Liberato, Khalil, and Habash}]{alhafni2024samer}
Bashar Alhafni, Reem Hazim, Juan~Pi{\~n}eros Liberato, Muhamed~Al Khalil, and Nizar Habash. 2024.
\newblock \href {https://arxiv.org/abs/2404.18615} {The samer arabic text simplification corpus}.
\newblock \emph{arXiv preprint arXiv:2404.18615}.

\bibitem[{Alhafni et~al.(2023)Alhafni, Inoue, Khairallah, and Habash}]{alhafni2023advancements}
Bashar Alhafni, Go~Inoue, Christian Khairallah, and Nizar Habash. 2023.
\newblock \href {https://arxiv.org/abs/2305.14734} {Advancements in arabic grammatical error detection and correction: An empirical investigation}.
\newblock \emph{arXiv preprint arXiv:2305.14734}.

\bibitem[{Belkebir and Habash(2021{\natexlab{a}})}]{belkebir2021automatic}
Riadh Belkebir and Nizar Habash. 2021{\natexlab{a}}.
\newblock \href {{https://arxiv.org/abs/2109.08068}} {Automatic error type annotation for arabic}.
\newblock \emph{arXiv preprint arXiv:2109.08068}.

\bibitem[{Belkebir and Habash(2021{\natexlab{b}})}]{Belkebir2021AutomaticET}
Riadh Belkebir and Nizar Habash. 2021{\natexlab{b}}.
\newblock \href {{https://arxiv.org/abs/2109.08068}} {Automatic error type annotation for arabic}.
\newblock In \emph{Conference on Computational Natural Language Learning}.

\bibitem[{Brown et~al.(2020)Brown, Mann, Ryder, Subbiah, Kaplan, Dhariwal, Neelakantan, Shyam, Sastry, Askell et~al.}]{brown2020language}
Tom Brown, Benjamin Mann, Nick Ryder, Melanie Subbiah, Jared~D Kaplan, Prafulla Dhariwal, Arvind Neelakantan, Pranav Shyam, Girish Sastry, Amanda Askell, et~al. 2020.
\newblock \href {{https://splab.sdu.edu.cn/GPT3.pdf}} {Language models are few-shot learners}.
\newblock \emph{Advances in neural information processing systems}, 33:1877--1901.

\bibitem[{Bryant et~al.(2017{\natexlab{a}})Bryant, Felice, and Briscoe}]{bryant-etal-2017-automatic}
Christopher Bryant, Mariano Felice, and Ted Briscoe. 2017{\natexlab{a}}.
\newblock \href {https://doi.org/10.18653/v1/P17-1074} {Automatic annotation and evaluation of error types for grammatical error correction}.
\newblock In \emph{Proceedings of the 55th Annual Meeting of the Association for Computational Linguistics (Volume 1: Long Papers)}, pages 793--805, Vancouver, Canada. Association for Computational Linguistics.

\bibitem[{Bryant et~al.(2017{\natexlab{b}})Bryant, Felice, and Briscoe}]{Bryant2017AutomaticAA}
Christopher Bryant, Mariano Felice, and Ted Briscoe. 2017{\natexlab{b}}.
\newblock \href {https://api.semanticscholar.org/CorpusID:12122749} {Automatic annotation and evaluation of error types for grammatical error correction}.
\newblock In \emph{Annual Meeting of the Association for Computational Linguistics}.

\bibitem[{Bubeck et~al.(2023)Bubeck, Chandrasekaran, Eldan, Gehrke, Horvitz, Kamar, Lee, Lee, Li, Lundberg et~al.}]{bubeck2023sparks}
S{\'e}bastien Bubeck, Varun Chandrasekaran, Ronen Eldan, Johannes Gehrke, Eric Horvitz, Ece Kamar, Peter Lee, Yin~Tat Lee, Yuanzhi Li, Scott Lundberg, et~al. 2023.
\newblock \href {{https://arxiv.org/abs/2303.12712}} {Sparks of artificial general intelligence: Early experiments with gpt-4}.
\newblock \emph{arXiv preprint arXiv:2303.12712}.

\bibitem[{Chamoun et~al.(2024)Chamoun, Schlichktrull, and Vlachos}]{chamoun2024automated}
Eric Chamoun, Michael Schlichktrull, and Andreas Vlachos. 2024.
\newblock \href {{https://arxiv.org/abs/2405.20477}} {Automated focused feedback generation for scientific writing assistance}.
\newblock \emph{arXiv preprint arXiv:2405.20477}.

\bibitem[{Chen et~al.(2024)Chen, Wang, Yu, and Zhou}]{article_chens}
Yanhan Chen, Hanxuan Wang, Kaiwen Yu, and Ruoshui Zhou. 2024.
\newblock \href {https://doi.org/10.54097/vfwgas09} {Artificial intelligence methods in natural language processing: A comprehensive review}.
\newblock \emph{Highlights in Science, Engineering and Technology}, 85:545--550.

\bibitem[{Chowdhery et~al.(2023)Chowdhery, Narang, Devlin, Bosma, Mishra, Roberts, Barham, Chung, Sutton, Gehrmann et~al.}]{chowdhery2023palm}
Aakanksha Chowdhery, Sharan Narang, Jacob Devlin, Maarten Bosma, Gaurav Mishra, Adam Roberts, Paul Barham, Hyung~Won Chung, Charles Sutton, Sebastian Gehrmann, et~al. 2023.
\newblock \href {{https://www.jmlr.org/papers/v24/22-1144.html}} {Palm: Scaling language modeling with pathways}.
\newblock \emph{Journal of Machine Learning Research}, 24(240):1--113.

\bibitem[{Chung et~al.(2024)Chung, Hou, Longpre, Zoph, Tay, Fedus, Li, Wang, Dehghani, Brahma et~al.}]{chung2024scaling}
Hyung~Won Chung, Le~Hou, Shayne Longpre, Barret Zoph, Yi~Tay, William Fedus, Yunxuan Li, Xuezhi Wang, Mostafa Dehghani, Siddhartha Brahma, et~al. 2024.
\newblock \href {{https://www.jmlr.org/papers/v25/23-0870.html}} {Scaling instruction-finetuned language models}.
\newblock \emph{Journal of Machine Learning Research}, 25(70):1--53.

\bibitem[{Du et~al.(2022)Du, Raheja, Kumar, Kim, Lopez, and Kang}]{du-etal-2022-understanding-iterative}
Wanyu Du, Vipul Raheja, Dhruv Kumar, Zae~Myung Kim, Melissa Lopez, and Dongyeop Kang. 2022.
\newblock \href {https://doi.org/10.18653/v1/2022.acl-long.250} {Understanding iterative revision from human-written text}.
\newblock In \emph{Proceedings of the 60th Annual Meeting of the Association for Computational Linguistics (Volume 1: Long Papers)}, pages 3573--3590, Dublin, Ireland. Association for Computational Linguistics.

\bibitem[{Einieh and AlMansour(2022)}]{Einieh_summarization}
Yasmin Einieh and Amal AlMansour. 2022.
\newblock \href {https://doi.org/10.1109/ESOLEC54569.2022.10009528} {Deep learning in arabic text summarization: Approaches, datasets, and evaluation metrics}.
\newblock In \emph{2022 20th International Conference on Language Engineering (ESOLEC)}, volume~20, pages 45--49.

\bibitem[{Fei et~al.(2023)Fei, Cui, Yang, Lam, Lan, and Shi}]{Fei2023EnhancingGE}
Yuejiao Fei, Leyang Cui, Sen Yang, Wai Lam, Zhenzhong Lan, and Shuming Shi. 2023.
\newblock \href {https://api.semanticscholar.org/CorpusID:258887567} {Enhancing grammatical error correction systems with explanations}.
\newblock \emph{ArXiv}, abs/2305.15676.

\bibitem[{Ferguson(2020)}]{Ferguson2020Diglossia}
Charles~A. Ferguson. 2020.
\newblock \href {https://api.semanticscholar.org/CorpusID:239352211} {Diglossia}.
\newblock \emph{The Bilingualism Reader}.

\bibitem[{Flower and Hayes(1980)}]{Flower}
Linda Flower and John~R. Hayes. 1980.
\newblock \href {http://www.jstor.org/stable/356630} {The cognition of discovery: Defining a rhetorical problem}.
\newblock \emph{College Composition and Communication}, 31(1):21--32.

\bibitem[{Habash and Palfreyman(2022)}]{habash-palfreyman-2022-zaebuc}
Nizar Habash and David Palfreyman. 2022.
\newblock \href {https://aclanthology.org/2022.lrec-1.9} {{ZAEBUC}: An annotated {A}rabic-{E}nglish bilingual writer corpus}.
\newblock In \emph{Proceedings of the Thirteenth Language Resources and Evaluation Conference}, pages 79--88, Marseille, France. European Language Resources Association.

\bibitem[{Hamani(2014)}]{hamani2014phenomenon}
Abdenbi Hamani. 2014.
\newblock \href {{https://tabayyun.dohainstitute.org/en/issue007/pages/art03.aspx}} {The phenomenon of i’rab: Between linguistic conventions and relations of proximity}.
\newblock \emph{Tabayyun}, 2(7):47--70.

\bibitem[{Hazim et~al.(2022)Hazim, Saddiki, Alhafni, Al~Khalil, and Habash}]{hazim-etal-2022-arabic}
Reem Hazim, Hind Saddiki, Bashar Alhafni, Muhamed Al~Khalil, and Nizar Habash. 2022.
\newblock \href {https://doi.org/10.18653/v1/2022.emnlp-demos.24} {{A}rabic word-level readability visualization for assisted text simplification}.
\newblock In \emph{Proceedings of the 2022 Conference on Empirical Methods in Natural Language Processing: System Demonstrations}, pages 242--249, Abu Dhabi, UAE. Association for Computational Linguistics.

\bibitem[{Holes(2004)}]{holes2004modern}
Clive Holes. 2004.
\newblock \href {{https://books.google.ae/books?hl=en&lr=&id=8E0Rr1xY4TQC&oi=fnd&pg=PR10&dq=Modern+Arabic:+Structures,+func-+tions,+and+varieties.+&ots=X3tFbjfhec&sig=FyTlcTQG12l8yTgSfEc1BTBQNRs&redir_esc=y#v=onepage&q=Modern%20Arabic%3A%20Structures%2C%20func-%20tions%2C%20and%20varieties.&f=false}} {\emph{Modern Arabic: Structures, functions, and varieties}}.
\newblock Georgetown University Press.

\bibitem[{Ippolito et~al.(2022)Ippolito, Yuan, Coenen, and Burnam}]{ippolito2022creative}
Daphne Ippolito, Ann Yuan, Andy Coenen, and Sehmon Burnam. 2022.
\newblock \href {{https://arxiv.org/abs/2211.05030}} {Creative writing with an ai-powered writing assistant: Perspectives from professional writers}.
\newblock \emph{arXiv preprint arXiv:2211.05030}.

\bibitem[{Kunz and Kuhlmann(2024)}]{kunz2024properties}
Jenny Kunz and Marco Kuhlmann. 2024.
\newblock \href {{https://arxiv.org/abs/2402.10532}} {Properties and challenges of llm-generated explanations}.
\newblock \emph{arXiv preprint arXiv:2402.10532}.

\bibitem[{Kwon et~al.(2023)Kwon, Bhatia, Nagoudi, and Abdul-Mageed}]{kwon2023beyond}
Sang~Yun Kwon, Gagan Bhatia, El~Moatez~Billah Nagoudi, and Muhammad Abdul-Mageed. 2023.
\newblock \href {{https://arxiv.org/abs/2312.08400}} {Beyond english: Evaluating llms for arabic grammatical error correction}.
\newblock \emph{arXiv preprint arXiv:2312.08400}.

\bibitem[{Lagrini and Redjimi(2021)}]{Lagrini2021ANA}
Samira Lagrini and Mohammed Redjimi. 2021.
\newblock \href {https://api.semanticscholar.org/CorpusID:245616869} {A new approach for arabic text summarization}.
\newblock In \emph{International Conference on Natural Language and Speech Processing}.

\bibitem[{Lee et~al.(2024)Lee, Gero, Chung, Shum, Raheja, Shen, Venugopalan, Wambsganss, Zhou, Alghamdi et~al.}]{lee2024design}
Mina Lee, Katy~Ilonka Gero, John Joon~Young Chung, Simon~Buckingham Shum, Vipul Raheja, Hua Shen, Subhashini Venugopalan, Thiemo Wambsganss, David Zhou, Emad~A Alghamdi, et~al. 2024.
\newblock \href {{https://arxiv.org/pdf/2403.14117}} {A design space for intelligent and interactive writing assistants}.
\newblock In \emph{Proceedings of the CHI Conference on Human Factors in Computing Systems}, pages 1--35.

\bibitem[{Li et~al.(2024)Li, Liang, Peng, and Yin}]{li2024value}
Zhuoyan Li, Chen Liang, Jing Peng, and Ming Yin. 2024.
\newblock \href {{https://dl.acm.org/doi/full/10.1145/3613904.3642625}} {The value, benefits, and concerns of generative ai-powered assistance in writing}.
\newblock In \emph{Proceedings of the CHI Conference on Human Factors in Computing Systems}, pages 1--25.

\bibitem[{Liu et~al.(2022)Liu, Tam, Muqeeth, Mohta, Huang, Bansal, and Raffel}]{liu2022few}
Haokun Liu, Derek Tam, Mohammed Muqeeth, Jay Mohta, Tenghao Huang, Mohit Bansal, and Colin~A Raffel. 2022.
\newblock \href {{https://proceedings.neurips.cc/paper_files/paper/2022/hash/0cde695b83bd186c1fd456302888454c-Abstract-Conference.html}} {Few-shot parameter-efficient fine-tuning is better and cheaper than in-context learning}.
\newblock \emph{Advances in Neural Information Processing Systems}, 35:1950--1965.

\bibitem[{Maamouri et~al.(2004)Maamouri, Bies, Buckwalter, and Mekki}]{maamouri2004penn}
Mohamed Maamouri, Ann Bies, Tim Buckwalter, and Wigdan Mekki. 2004.
\newblock \href {{https://www.marefa.org/images/e/e8/The_penn_arabic_treebank_Building_a_large-scale_an_%281%29.pdf}} {The penn arabic treebank: Building a large-scale annotated arabic corpus}.
\newblock In \emph{NEMLAR conference on Arabic language resources and tools}, volume~27, pages 466--467. Cairo.

\bibitem[{Mohamed et~al.(2022)Mohamed, Khelil, Savary, Keskes, Antoine, and Hadrich}]{mohamed2022annotating}
Najet~Hadj Mohamed, Cherifa~Ben Khelil, Agata Savary, Iskandar Keskes, Jean-Yves Antoine, and Lamia~Belguith Hadrich. 2022.
\newblock \href {{https://hal.science/hal-03712937/}} {Annotating verbal multiword expressions in arabic: Assessing the validity of a multilingual annotation procedure}.
\newblock In \emph{13th Conference on Language Resources and Evaluation (LREC 2022)}, pages 1839--1848.

\bibitem[{Mostefa et~al.(2015)Mostefa, Abualasal, Asbayou, Gzawi, and Abb{\`e}s}]{Mostefa2015TECHLIMEDQALBSharedT2}
Djamel Mostefa, Jaber Abualasal, Omar Asbayou, Mahmoud Gzawi, and Ramzi Abb{\`e}s. 2015.
\newblock \href {https://api.semanticscholar.org/CorpusID:17540112} {Techlimed@qalb-shared task 2015: a hybrid arabic error correction system}.
\newblock In \emph{ANLP@ACL}.

\bibitem[{Ouyang et~al.(2022)Ouyang, Wu, Jiang, Almeida, Wainwright, Mishkin, Zhang, Agarwal, Slama, Ray et~al.}]{ouyang2022training}
Long Ouyang, Jeffrey Wu, Xu~Jiang, Diogo Almeida, Carroll Wainwright, Pamela Mishkin, Chong Zhang, Sandhini Agarwal, Katarina Slama, Alex Ray, et~al. 2022.
\newblock \href {{https://proceedings.neurips.cc/paper_files/paper/2022/hash/b1efde53be364a73914f58805a001731-Abstract-Conference.html}} {Training language models to follow instructions with human feedback}.
\newblock \emph{Advances in Neural Information Processing Systems}, 35:27730--27744.

\bibitem[{Peng et~al.(2023)Peng, Li, He, Galley, and Gao}]{peng2023instruction}
Baolin Peng, Chunyuan Li, Pengcheng He, Michel Galley, and Jianfeng Gao. 2023.
\newblock \href {{https://arxiv.org/abs/2304.03277}} {Instruction tuning with gpt-4}.
\newblock \emph{arXiv preprint arXiv:2304.03277}.

\bibitem[{Raheja et~al.(2023)Raheja, Kumar, Koo, and Kang}]{raheja2023coedit}
Vipul Raheja, Dhruv Kumar, Ryan Koo, and Dongyeop Kang. 2023.
\newblock \href {{https://arxiv.org/abs/2305.09857}} {Coedit: Text editing by task-specific instruction tuning}.
\newblock \emph{arXiv preprint arXiv:2305.09857}.

\bibitem[{Reid et~al.(2024)Reid, Savinov, Teplyashin, Lepikhin, Lillicrap, Alayrac, Soricut, Lazaridou, Firat, Schrittwieser et~al.}]{reid2024gemini}
Machel Reid, Nikolay Savinov, Denis Teplyashin, Dmitry Lepikhin, Timothy Lillicrap, Jean-baptiste Alayrac, Radu Soricut, Angeliki Lazaridou, Orhan Firat, Julian Schrittwieser, et~al. 2024.
\newblock \href {{https://arxiv.org/abs/2403.05530}} {Gemini 1.5: Unlocking multimodal understanding across millions of tokens of context}.
\newblock \emph{arXiv preprint arXiv:2403.05530}.

\bibitem[{Rozovskaya et~al.(2015)Rozovskaya, Bouamor, Habash, Zaghouani, Obeid, and Mohit}]{rozovskaya2015second}
Alla Rozovskaya, Houda Bouamor, Nizar Habash, Wajdi Zaghouani, Ossama Obeid, and Behrang Mohit. 2015.
\newblock \href {{https://aclanthology.org/W15-3204.pdf}} {The second qalb shared task on automatic text correction for arabic}.
\newblock In \emph{Proceedings of the Second workshop on Arabic natural language processing}, pages 26--35.

\bibitem[{Rozovskaya et~al.(2014)Rozovskaya, Habash, Eskander, Farra, and Salloum}]{rozovskaya2014columbia}
Alla Rozovskaya, Nizar Habash, Ramy Eskander, Noura Farra, and Wael Salloum. 2014.
\newblock \href {{https://aclanthology.org/W14-3622.pdf}} {The columbia system in the qalb-2014 shared task on arabic error correction}.
\newblock In \emph{Proceedings of the EMNLP 2014 Workshop on Arabic Natural Language Processing (ANLP)}, pages 160--164.

\bibitem[{Saddiki et~al.(2018)Saddiki, Habash, Cavalli-Sforza, and Al~Khalil}]{saddiki-etal-2018-feature}
Hind Saddiki, Nizar Habash, Violetta Cavalli-Sforza, and Muhamed Al~Khalil. 2018.
\newblock \href {https://doi.org/10.18653/v1/W18-3703} {Feature optimization for predicting readability of {A}rabic {L}1 and {L}2}.
\newblock In \emph{Proceedings of the 5th Workshop on Natural Language Processing Techniques for Educational Applications}, pages 20--29, Melbourne, Australia. Association for Computational Linguistics.

\bibitem[{Sag et~al.(2002)Sag, Baldwin, Bond, Copestake, and Flickinger}]{sag2002multiword}
Ivan~A Sag, Timothy Baldwin, Francis Bond, Ann Copestake, and Dan Flickinger. 2002.
\newblock \href {{https://link.springer.com/chapter/10.1007/3-540-45715-1_1}} {Multiword expressions: A pain in the neck for nlp}.
\newblock In \emph{Computational Linguistics and Intelligent Text Processing: Third International Conference, CICLing 2002 Mexico City, Mexico, February 17--23, 2002 Proceedings 3}, pages 1--15. Springer.

\bibitem[{Sanh et~al.(2021)Sanh, Webson, Raffel, Bach, Sutawika, Alyafeai, Chaffin, Stiegler, Scao, Raja et~al.}]{sanh2021multitask}
Victor Sanh, Albert Webson, Colin Raffel, Stephen~H Bach, Lintang Sutawika, Zaid Alyafeai, Antoine Chaffin, Arnaud Stiegler, Teven~Le Scao, Arun Raja, et~al. 2021.
\newblock Multitask prompted training enables zero-shot task generalization.
\newblock \emph{arXiv preprint arXiv:2110.08207}.

\bibitem[{Sawalha et~al.(2013)Sawalha, Atwell, and Abushariah}]{Sawalha2013SALMASA}
Majdi Sawalha, Eric Atwell, and Mohammad Abd-Alrahman~Mahmoud Abushariah. 2013.
\newblock \href {https://api.semanticscholar.org/CorpusID:14879997} {Salma: Standard arabic language morphological analysis}.
\newblock \emph{2013 1st International Conference on Communications, Signal Processing, and their Applications (ICCSPA)}, pages 1--6.

\bibitem[{Schick et~al.(2022)Schick, Dwivedi-Yu, Jiang, Petroni, Lewis, Izacard, You, Nalmpantis, Grave, and Riedel}]{schick2022peer}
Timo Schick, Jane Dwivedi-Yu, Zhengbao Jiang, Fabio Petroni, Patrick Lewis, Gautier Izacard, Qingfei You, Christoforos Nalmpantis, Edouard Grave, and Sebastian Riedel. 2022.
\newblock \href {{https://arxiv.org/abs/2208.11663}} {Peer: A collaborative language model}.
\newblock \emph{arXiv preprint arXiv:2208.11663}.

\bibitem[{Solyman et~al.(2019)Solyman, Wang, and Tao}]{Solyman2019ProposedMF}
Aiman Solyman, Zhenyu Wang, and Qian Tao. 2019.
\newblock \href {https://api.semanticscholar.org/CorpusID:216043920} {Proposed model for arabic grammar error correction based on convolutional neural network}.
\newblock \emph{2019 International Conference on Computer, Control, Electrical, and Electronics Engineering (ICCCEEE)}, pages 1--6.

\bibitem[{Solyman et~al.(2021)Solyman, Zhenyu, Qian, Elhag, Toseef, and Aleibeid}]{SOLYMAN2021303}
Aiman Solyman, Wang Zhenyu, Tao Qian, Arafat Abdulgader~Mohammed Elhag, Muhammad Toseef, and Zeinab Aleibeid. 2021.
\newblock \href {https://doi.org/https://doi.org/10.1016/j.eij.2020.12.001} {Synthetic data with neural machine translation for automatic correction in arabic grammar}.
\newblock \emph{Egyptian Informatics Journal}, 22(3):303--315.

\bibitem[{Song et~al.(2023)Song, Krishna, Bhatt, Gimpel, and Iyyer}]{song2023gee}
Yixiao Song, Kalpesh Krishna, Rajesh Bhatt, Kevin Gimpel, and Mohit Iyyer. 2023.
\newblock \href {{https://arxiv.org/abs/2311.09517}} {Gee! grammar error explanation with large language models}.
\newblock \emph{arXiv preprint arXiv:2311.09517}.

\bibitem[{Urmeneta and Romero(2024)}]{book_woo}
Alex Urmeneta and Margarida Romero. 2024.
\newblock \href {https://doi.org/10.1007/978-3-031-55272-4} {\emph{Creative Applications of Artificial Intelligence in Education}}.

\bibitem[{Versteegh(2014)}]{versteegh2014arabic}
Kees Versteegh. 2014.
\newblock \href {{https://books.google.ae/books?hl=en&lr=&id=RiarBgAAQBAJ&oi=fnd&pg=PR3&dq=Kees+Versteegh.+2014.+Arabic+language.&ots=9iqhDdj1g9&sig=0s_G_eIjcSG9EVelWEEUs1tQ9fk&redir_esc=y#v=onepage&q=Kees%20Versteegh.%202014.%20Arabic%20language.&f=false}} {\emph{Arabic language}}.
\newblock Edinburgh University Press.

\bibitem[{Wei et~al.(2022)Wei, Bosma, Zhao, Guu, Yu, Lester, Du, Dai, and Le}]{wei2022finetuned}
Jason Wei, Maarten Bosma, Vincent Zhao, Kelvin Guu, Adams~Wei Yu, Brian Lester, Nan Du, Andrew~M. Dai, and Quoc~V Le. 2022.
\newblock \href {https://openreview.net/forum?id=gEZrGCozdqR} {Finetuned language models are zero-shot learners}.
\newblock In \emph{International Conference on Learning Representations}.

\bibitem[{White et~al.(2023)White, Fu, Hays, Sandborn, Olea, Gilbert, Elnashar, Spencer-Smith, and Schmidt}]{white2023prompt}
Jules White, Quchen Fu, Sam Hays, Michael Sandborn, Carlos Olea, Henry Gilbert, Ashraf Elnashar, Jesse Spencer-Smith, and Douglas~C Schmidt. 2023.
\newblock \href {{https://arxiv.org/abs/2302.11382}} {A prompt pattern catalog to enhance prompt engineering with chatgpt}.
\newblock \emph{arXiv preprint arXiv:2302.11382}.

\bibitem[{Zaghouani et~al.(2014)Zaghouani, Mohit, Habash, Obeid, Tomeh, Rozovskaya, Farra, Alkuhlani, and Oflazer}]{zaghouani-etal-2014-large}
Wajdi Zaghouani, Behrang Mohit, Nizar Habash, Ossama Obeid, Nadi Tomeh, Alla Rozovskaya, Noura Farra, Sarah Alkuhlani, and Kemal Oflazer. 2014.
\newblock \href {http://www.lrec-conf.org/proceedings/lrec2014/pdf/956_Paper.pdf} {Large scale {A}rabic error annotation: Guidelines and framework}.
\newblock In \emph{Proceedings of the Ninth International Conference on Language Resources and Evaluation ({LREC}'14)}, Reykjavik, Iceland. European Language Resources Association (ELRA).

\bibitem[{Zhang et~al.(2023)Zhang, Dong, Li, Zhang, Sun, Wang, Li, Hu, Zhang, Wu et~al.}]{zhang2023instruction}
Shengyu Zhang, Linfeng Dong, Xiaoya Li, Sen Zhang, Xiaofei Sun, Shuhe Wang, Jiwei Li, Runyi Hu, Tianwei Zhang, Fei Wu, et~al. 2023.
\newblock \href {{https://arxiv.org/abs/2308.10792}} {Instruction tuning for large language models: A survey}.
\newblock \emph{arXiv preprint arXiv:2308.10792}.

\bibitem[{Zhang et~al.(2022)Zhang, Roller, Goyal, Artetxe, Chen, Chen, Dewan, Diab, Li, Lin et~al.}]{zhang2022opt}
Susan Zhang, Stephen Roller, Naman Goyal, Mikel Artetxe, Moya Chen, Shuohui Chen, Christopher Dewan, Mona Diab, Xian Li, Xi~Victoria Lin, et~al. 2022.
\newblock \href {{https://arxiv.org/abs/2205.01068}} {Opt: Open pre-trained transformer language models}.
\newblock \emph{arXiv preprint arXiv:2205.01068}.

\end{thebibliography}
\bibliographystyle{acl_natbib}

\appendix

\clearpage
\appendixpage
\addappheadtotoc
\numberwithin{figure}{section}
\numberwithin{table}{section}

We provide an addition organized as follows:

\begin{itemize}
\item Detailed Related Works \ref{apndx:related_works}.
\item Linguistic Background of Arabic \ref{apdx:linguistic_background_of_arabic}.
\item Extended ALC Error Taxonomy \ref{extended_acl_errors_appdx}.
\item ATB Synthetic Data \ref{apdx:synthetic_data}.
\item Prompt Design \ref{apdx:prompt_design}.
\item Evaluation matrices for Arabic Writing Assistant Tasks \ref{apdx:matrices}.

\end{itemize}

\section{Detailed Related Works}
\label{apndx:related_works}
\paragraph{LLM’s for Writing assistance.}
Advancements in instruction tuning for writing assistance include works like PEER~\cite{schick2022peer}, which enhances editing by adhering to and explaining a user-provided plan, and CoEdIT~\cite{raheja2023coedit}, which focuses on managing editing plans to refine and clarify content without introducing new information. Additionally, SWIF\textsuperscript{2}T~\cite{chamoun2024automated}, designed for scientific writing, generates specific, actionable comments to identify weaknesses in a paper and suggest revisions. These developments highlight the progress in creating sophisticated tools that enhance the writing and editing processes across various applications.

\paragraph{Progress in Arabic Writing Tasks.}
In GEC, advances include the QALB corpus used for the QALB 2014 and 2015 shared tasks~\cite{zaghouani-etal-2014-large,rozovskaya2015second}, and the Zaebuc corpus~\cite{habash-palfreyman-2022-zaebuc}. Advances in model techniques include Convolutional Neural Networks, alignment algorithms, LLMs~\cite{Solyman2019ProposedMF, SOLYMAN2021303, alhafni2023advancements, kwon2023beyond}. In text simplification, tools such as visualization for assisted text simplification~\cite{hazim-etal-2022-arabic} and feature optimization techniques~\cite{saddiki-etal-2018-feature} have been developed, along with corpus advancements like the SAMER corpus~\cite{al-khalil-etal-2018-leveled, alhafni2024samer}. For text summarization, model developments include survey approaches~\cite{Einieh_summarization}, rhetorical structure theory with statistical methods~\cite{Lagrini2021ANA}, and deep learning approaches~\cite{Al-Maleh_summarization}.

\paragraph{Explainable Writing Systems.}
Explainable writing systems enhance the transparency of language processing tools by providing clear and understandable insights aligned with instructional goals. For example, ERRANT~\cite{Bryant2017AutomaticAA} annotates grammatical errors with clear edit boundaries and error types. EXPECT~\cite{Fei2023EnhancingGE} offers a dataset with detailed annotations to support explainable grammatical error correction. GEE~\cite{song2023gee} advances this by providing one-sentence explanations for grammatical errors in pairs of erroneous and corrected sentences. In text simplification and feedback generation, systems strive to provide clear and relevant explanations~\cite{agarwal2022explain}. These efforts highlight the importance of creating tools that deliver understandable insights and ensure alignment with instructional goals. Additionally, understanding the properties and challenges of LLM-generated explanations~\cite{kunz2024properties} is crucial for maintaining user trust and comprehension.

\section{Linguistic Background of Arabic}
\label{apdx:linguistic_background_of_arabic}
Arabic is a highly sophisticated and complex language, distinguished by the richness of its morphological and grammatical features~\cite{holes2004modern}. It is linguistically diverse, encompassing numerous dialects across a widespread geographical region~\cite{versteegh2014arabic}. Arabic is recognized as a diglossic language, where both Modern Standard Arabic (MSA) and various dialects coexist~\cite{Ferguson2020Diglossia}. MSA is employed in formal contexts, whereas local dialects dominate everyday communication, exhibiting significant differences in pronunciation, vocabulary, and grammar.
The inherent complexity and diversity of Arabic presents considerable challenges in the field of NLP, making efforts in writing tasks and writing assistance valuable. For instance, Arabic verbs are intricate, changing based on tense, mood, voice, and gender, posing significant challenges for writing tasks. Moreover, the grammar includes complex rules, such as changes in noun forms depending on their grammatical role. Additionally, Arabic is orthographically ambiguous due to its reliance on consonants and the optional use of diacritics to denote short vowels and other phonetic details~\cite{Sawalha2013SALMASA}. This ambiguity leads to multiple interpretations of the same consonant string without context or diacritics, complicating writing in Arabic~\cite{abdul-mageed-etal-2020-nadi, belkebir2021automatic}. For example, the word \AR{\small{علم}} with diacritics has three different meanings: \AR{\small{عَلِمَ}} meaning (he knew), \AR{\small{عِلْم}} meaning (knowledge), and \AR{\small{عَلَم}} meaning (flag).

Furthermore, the morphological ambiguity of Arabic occurs when the same root or pattern leads to different meanings based on context, inflection, or derivation. Arabic's rich system of roots and patterns (templates) allows for a single root to generate various words with different meanings and grammatical roles. The root of the verb \AR{\small{ض-ر-ب}} (hit), for instance, has different patterns indicating tense, voice (active/passive), and other grammatical nuances, such as \AR{\small{ضَرَبَ}} (daraba - he hit) vs. \AR{\small{\RL{يُضْرِب}}} (yudrib - he is hitting).

\section{Writing Assistance Instructions}
\label{writing_assistance_instructions}
\paragraph{GEC Instructions.}
For example, the Orthographic error class features a subclass for \textit{hamza} errors, further divided into four sub-subclasses: \textit{medial hamza}, \textit{final hamza}, \textit{cutting hamza}, and \textit{hamzet al-wasl}. Our instructions address a broad spectrum of these sub-classes and sub-subclasses, ensuring comprehensive coverage across different error categories.

\paragraph{GEC Error Explanations.}
In our dataset, we provide explanation of the GEC errors in both Arabic and English. For instance, consider the sentence \AR{\small{يجري ضد الحقوق الإنسان}} (goes against human rights), which contains a \textit{Syntax} error. The English explanation is: Adding the definite article \AR{\small{ال}} to the annexed noun is incorrect because the annexed noun and the annexing noun should share the same definiteness or indefiniteness. In another example, \AR{\small{ شعرت بالسعادة عندما وطأت أرض الوطن.}}, which contains an \textit{Orthographic} error, the Arabic explanation is: \AR{\small{وطئت: كتبت الهمزة المتوسطة على ياء لأن حركتها السكون وحركة الحرف الذي قبلها الكسرة، والكسرة أقوى من السكون وتناسبها الياء، لذلك كتبت على ياء.}}

\paragraph{ATB Synthetic Data.}
We utilize all parts of the ATB to generate 100 examples related to different classes of errors in our taxonomy, covering the following error classes and sub-classes:

\begin{itemize}
    \item \textbf{Orthographic} errors, including:
\end{itemize}
i) \textit{Wrong Order of Word Characters (OC)},
ii) \textit{Replacement in Word Characters (OR)},
iii) \textit{Additional Characters (OD)},
iv) \textit{Missing Characters (OM)}.

\begin{itemize}
    \item \textbf{Syntax} errors, which include:
\end{itemize}
i) \textit{Missing Words (XM)},
ii) \textit{Unnecessary Words (XT)}.

\begin{itemize}
    \item \textbf{Morphological} errors, involving:
i) \textit{Word Inflection (MI)},
ii) \textit{Verb Tense (MT)}.
\end{itemize}
\begin{itemize}
    \item \textbf{Split} and \textbf{Merge}
\end{itemize}

\begin{itemize}
    \item \textbf{Punctuation} errors, categorized into:
\end{itemize}
i) \textit{Punctuation Confusion (PC)},
ii) \textit{Missing Punctuation (PM)},
iii) \textit{Unnecessary Punctuation (PT)},
iv) \textit{Other Punctuation Errors (PO)}.

\begin{itemize}
    \item \textbf{Text Refinement}
\end{itemize}
For instance, the informal use of the following expression in MSA:    
\AR{\small{نعم، يا سعد بيه، ليس فيها أي خدوش، موجودة، وتممت عليها في مخازن الشركة.}} 
would be polished to be:
\AR{\small{نعم، يا سيد سعد، ليس فيها أي خدوش، موجودة, وتممت عليها في مخازن الشركة.}}.The explanation provided is: The word \AR{\small{بيه}}  is an Egyptian title indicating respect and does not exist in Standard Arabic, so it was necessary to phrase it as \AR{\small{سيد}}. Another example of text refinement is \AR{\small{أنتم ما زلتم جالسن ترغون }} and the polished version of it is: \AR{\small{أما زلتما تثرثران؟}}. The explanation provided: \AR{\small{ترغون}} is an incorrect word and does not exist in Standard Arabic; the correct equivalent is \AR{\small{الثرثرة}}.

\section{Extended ALC Error Taxonomy}
\label{extended_acl_errors_appdx}

\textbf{Orthographic Errors} refer to  mistakes in the spelling of a word. It occurs when the arrangement of letters does not match the standard or accepted spelling. These errors can happen for several reasons, such as typing mistakes, incorrect memory of the spelling, or unfamiliarity with the spelling rules of a language. Orthographic errors are particularly common in languages with complex and irregular spelling rules, like Arabic. For instance, there are common spelling errors in \textit{medial hamza} as in \AR{\small{راى}} instead of \AR{\small{رأى}}. Other error examples include the confusion between \textit{Ta mabsouta} and \textit{Ta marbouta} as in \AR{\small{طارة}} instead of \AR{\small{طارت}}. Within the Orthographic errors category, which originally encompasses 12 subclasses as per the ALC framework, we have introduced additional granularity by defining further sub-subclasses. These include four distinct categories for Hamza errors: i) \textit{Medial hamza}, ii) \textit{Final hamza}, iii) \textit{Cutting hamza}, and iv) \textit{Hamzat al-wasl}.
 Additionally, we address the common confusions between Ha, Ta, and Ta marbouta by subdividing them into i) \textit{Ta mabsouta}, ii) \textit{Ta marbouta}, and iii) \textit{Ha}.\\
 \textbf{Syntax Errors} in language use refer to errors in the structure of a sentence which include the arrangement of words and phrases in ways that do not conform to the grammatical rules of the language, making the sentence unclear or incorrect. Syntax errors can involve incorrect word order, missing components like subjects or verbs, or improper use of grammatical forms such as agreement, or conjugation. In Arabic, errors include the gender, number, cases, etc. For example, \AR{\small{أطلق الجنديُّ رصاصتان}} the error in dual word \AR{\small{رصاصتان}} which should be \AR{\small{صاصتَينِ}} in Arabic. 
 For the Syntax error category, which initially comprised seven subclasses, we have introduced an additional subclass titled \textit{Closed class errors}. Further, the gender error subclass is now refined into a sub-subclass  errors of using feminine and masculine forms and vice versa. The number subclass has been divided into four sub-subclasses: i) \textit{singular}, ii) \textit{dual}, iii) \textit{plural}, and iv) \textit{numbers} with additional subdivisions under plural—namely i) \textit{sound masculine plural}, ii) \textit{sound feminine plural}, iii) \textit{broken plural}, and iv) \textit{Al-maqsour plural}. Additionally, we have structured the number into three further sub-subclasses: i) \textit{singular numbers}, ii) \textit{composite numbers}, and iii) \textit{Al-Aqoud}.\\
 
 We further refined the taxonomy by introducing two sub-subclasses under definiteness, specifically  addition and omission of the definite article "Al."
 Additionally, we have delineated five sub-subclasses within the category of case errors: i) \textit{nominative}, ii) \textit{genitive}, iii) \textit{accusative}, iv) \textit{jussive}, v) and \textit{other case errors}. We also introduced a new subclass titled \textit{Closed class category}. This category captures miscellaneous grammatical inaccuracies involving i) \textit{pronouns}, ii) \textit{the five verbs}, and iii) \textit{the five nouns} that do not fit neatly into other predefined categories. The pronouns sub-subclass is divided into three specific issues: i) \textit{adding unnecessary pronoun}, ii) \textit{removing necessary pronoun}, iii) and \textit{wrong pronoun reference}.\\

\begin{figure*}[!htp]
    \centering
    \includegraphics[width=\linewidth]{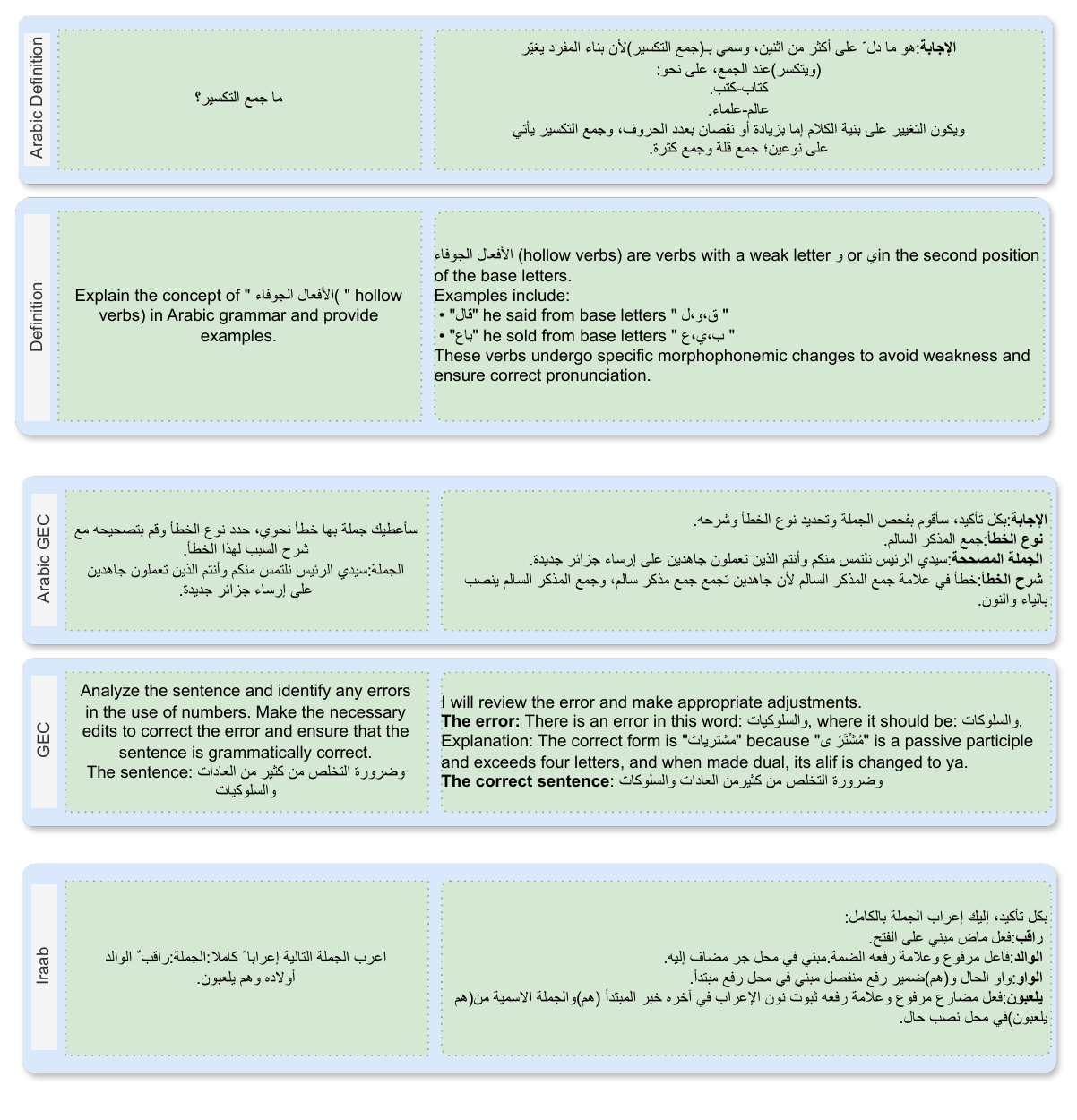}
    \caption{Examples of writing assistance instructions in Arabic and English.} 
    \label{fig:subtasks}
\end{figure*}

\begin{table*}[]
\resizebox{\textwidth}{!}{%
\begin{tabular}{ccllllr}
\toprule

\textbf{Class} &
  \textbf{Tag} &
  \textbf{Subclass} &
  \multicolumn{1}{l}{\textbf{Sub2}} &
  \multicolumn{1}{l}{\textbf{Sub3}} &
  \multicolumn{1}{l}{\textbf{Description}} &
  \multicolumn{1}{r}{\textbf{Example}} \\ \toprule
   &
   &
  \multicolumn{1}{l}{} &
  Medial Hamza &
  --- &
  Medial Hamza E. &
  \AR{{كتاءب ← كتائب}} 
   \\ \cmidrule{4-7}
   
 &
 OH
   &
  \multicolumn{1}{l}{} &
  Final Hamza &
  --- &
  Final Hamza E. &
  \AR{{تباطو ← تباطؤ}} 
   \\ \cmidrule{4-7}
 &
   &
  \multicolumn{1}{l}{} &
  Cutting Hamzah &
  --- &
  Cutting Hamzah E. &
  \AR{{اسهمت  ← أسهمت}} 
   \\ \cmidrule{4-7}
  
 &
  \multirow{-4}{*}{} &
  \multicolumn{1}{l}{\multirow{-4}{*}{Hamza}} &
  Hamzet al-Wasl &
  --- &
  Hamzet al Wasel E. &
  \AR{{إستعلمت ← استعلمت}} 
   \\  \cmidrule{2-7}

 &
  OA &
  \multicolumn{1}{l}{\begin{tabular}[c]{@{}l@{}}C. in  Alif\\ Ya and  Alif-Maqsura\end{tabular}} &
  Shortening Alif/ Alif Maqsoura &
  --- &
  C. in Alif Ya, and Alif-Maqsura &
  \AR{{مستوي ← مستوى}} 
   \\  \cmidrule{2-7}
 &
  OW &
  \multicolumn{1}{l}{C. in Alif  Fariqa} &
   &
  --- &
  C. in Alif Fariqa & 
  \RL{یسمُوا ← یسمُو} 
   \\ \cmidrule{4-7}
 &
   &
  \multicolumn{1}{l}{} &
  Ta Mabsouta &
  --- &
  C. in Ta Mabsouta &
  \AR{{طارة ← طارت} } 
   \\ \cmidrule{2-7}
 &
   &
  \multicolumn{1}{l}{} &
  Ta Marbouta &
  --- &
  C. in Ta Marbouta &
  \AR{{أتنزة ← أتنزه }} 
   \\
 &
  \multirow{-3}{*}{OT} &
  \multicolumn{1}{l}{\multirow{-3}{*}{C. in Ha, Ta and Ta-Marbuta}} &
  Ha &
  --- &
  C. in Ha &
  \AR{{كتابة ← كتابه} } 
   \\  \cmidrule{2-7}
   &
  ON &
  \multicolumn{1}{l}{C. between Nun and Tanwin} &
  \multicolumn{1}{l}{---} &
  --- &
  C. between Nun and Tanwin &
  \RL{{مثابرن ← مثابرٌ} } 
   \\ \cmidrule{3-7}
 &
  OS &
  \multicolumn{1}{l}{Shortening long vowels} &
  \multicolumn{1}{l}{---} &
  --- &
  Shortening long vowels &
  \AR{{ذكرتهم ← ذاكرتهم }} 
   \\ \cmidrule{2-7}
 &
  OG &
  \multicolumn{1}{l}{Lengthening short vowels} &
  \multicolumn{1}{l}{---} &
  --- &
  Lengthening short vowels &
 \AR{{ هاكذا ← هكذا} } 
   \\ \cmidrule{2-7}
 &
  OC &
  \multicolumn{1}{l}{Wrong order of word Chr.(s)} &
  \multicolumn{1}{l}{---} &
  --- &
  Wrong order of word Chr.(s) &
 \AR{{ كمب ← بكم} } 
   \\ \cmidrule{2-7}
   
 &
  OR &
  \multicolumn{1}{l}{Replacement in Word Chr.(s)} &
  \multicolumn{1}{l}{---} &
  --- &
Rep. in word Chr.(s) &
  \AR{{تحوبل ← تحويل}} 
   \\ \cmidrule{2-7}
   
 &
  OD &
  \multicolumn{1}{l}{Additional Chr.(s)} &
  \multicolumn{1}{l}{---} &
  ---&
  Adding unnecessary Chr.(s) &
 \AR{{ اطسم ← اسم} } 
   \\ \cmidrule{2-7}

 &
  OM &
  \multicolumn{1}{l}{Missing Chr.(s)} &
  \multicolumn{1}{l}{---} &
  --- &
  Missing necessary Chr.(s) &
  \AR{{ قنال ← قنابل} } 
   \\ \cmidrule{2-7}
\multirow{-17}{*}{\rotatebox[origin=c]{90}{\textbf{\colorbox{blue!10}{Orthography}}}} &
  OO &
  \multicolumn{1}{l}{Other Orthographic E.} &
  \multicolumn{1}{l}{---} &
  --- &
  Other orthographic E. &
  \multicolumn{1}{r}{---} 
  \\ \cmidrule{1-7}
  &
   &
   &
  Fem. Vs. Masc. &
  --- &
  C. in Fem. Vs. Masc. &
  \AR{{يشرب ← تشرب}} 
   \\ \cmidrule{4-7}
 & XG &
  Gender 
   &
  Singular &
  --- &
  C. in Sing. Vs. Pl. &
 \AR{{ النحل ← النحلة }} 
   \\ \cmidrule{4-7}

 &
   &
   &
  Dual &
  --- &
  C. in Dual &
\AR{{  المعلمين ← المعلمان} } 
   \\\cmidrule{2-7}
 &
   &
   &
  \multicolumn{1}{c}{} &
  Sound Masc. Pl. &
  Wrong Sound Masc. Pl. &
 \AR{{ مؤدب ← مؤدبون} } 
   \\ \cmidrule{5-7}

 &
   &
   &
  \multicolumn{1}{c}{} &
  Broken Pl. &
  Wrong Broken Pl. &
\AR{{  سعيدين ← سعداء }} 
   \\ \cmidrule{5-7}

 &
   &
   &
  \multicolumn{1}{c}{} &
  Almaqsour Pl. &
  Wrong AlMaqsour Pl. &
 \AR{{ المشتروات ← المشتريات }} 
   \\ \cmidrule{5-7}
 &
   &
   &
  \multicolumn{1}{l}{\multirow{-4}{*}{Plural}} &
  Sound Fem. Pl. &
  Wrong Sound Fem. Pl. &
 \AR{{  كتبوا ← كتبن} } 
   \\ \cmidrule{4-7}

    &
   &
   &
  \multicolumn{1}{c}{} &
  Sing. Num. &
  C. in Sing. Num. &
 \AR{{ ثلاث أيام ← ثلاثة أيام}} 
   \\ \cmidrule{5-7}
         
 &
   &
   &
  \multicolumn{1}{c}{} &
  Composite Num. &
  C. in Composite Num. &
 \AR{{ أربع عشر مبدعا ← أربعة عشر }} 
   \\ \cmidrule{5-7}
 &
  \multirow{-9}{*}{XN} &
  \multirow{-9}{*}{Number} &
  \multicolumn{1}{l}{\multirow{-3}{*}{Numbers}} &
  Al-Aqoud &
  C. in Al-Aqoud &
 \AR{{ الخمسون ← الخمسين }} 
   \\ \cmidrule{2-7}

 &
   &
   &
  Adding ``Al'' &
  --- &
   Adding ``Al'' &
 \AR{{ المنتصف ←  منتصف} } 
   \\ \cmidrule{4-7}
 &
  \multirow{-2}{*}{XF} &
  \multirow{-2}{*}{Definiteness} &
  Removing ``Al'' &
  --- &
  C. in Removing Al &
 \AR{{ عالمي ← العالمي} } 
   \\ \cmidrule{2-7}
 &
   &
   &
  Nom. Case E. &
  --- &
  C. in Nom. Case &
 \AR{{ الممرضين ← الممرضون} } 
   \\ \cmidrule{4-7}
 &
   &
   &
  Gen. Case E. &
  --- &
  C. in Gen. Case &
\AR{{  على عرفاتً ← عرفاتٍ }} 
   \\ \cmidrule{4-7}
 &
   &
   &
  Acc. Case E. &
  --- &
  C. in Acc. Case &
\AR{{ عادي ← عاديا }} 
   \\ \cmidrule{4-7}
 &
   &
   &
Juss. Case E. &
  --- &
  C. in Juss. Case &
\AR{{ يدعوا ← يدعُ }} 
   \\ \cmidrule{4-7}

 &
  \multirow{-5}{*}{XC} &
  \multirow{-5}{*}{Case} &
  Other Case E. &
  --- &
  Other Case E. &
  \multicolumn{1}{c}{-} 
   \\ \cmidrule{2-7}
 
 &
   &
  \multicolumn{1}{l}{} &
  \multicolumn{1}{c}{} &
  Adding Unnecessary Pron. &
  C. in Adding Pron. &
 \AR{{ أنا رأيته ← رأيت }} 
   \\ \cmidrule{5-7}
 &
   &
  \multicolumn{1}{l}{} &
  \multicolumn{1}{c}{} &
  Removing Necessary Pron. &
  C. in Removing Pron. &
 \AR{{ أخذ  ← أخذنا }} 
   \\ \cmidrule{5-7}
 &
   &
  \multicolumn{1}{l}{} &
  \multicolumn{1}{l}{\multirow{-3}{*}{Pronouns}} &
  Wrong Pron. Ref. &
  C. in Pron. Ref. &
 \AR{{ منازلهم ← منازلهن }} 
   \\ \cmidrule{4-7}
 &
   &
  \multicolumn{1}{l}{} &
  The Five N. &
  --- &
  C. in Five N. &
\AR{{  أبوك ←  أبيك }} 
   \\ \cmidrule{4-7}
 &
  \multirow{-5}{*}{XR} &
  \multicolumn{1}{l}{\multirow{-5}{*}{Closed Class E.}} &
  The Five V. &
  --- &
  C. in Five V. &
 \AR{{ تقنعى ←  تقنعين }} 
   \\ \cmidrule{2-7}
 &
  XM &
  \multicolumn{1}{l}{Missing Word} &
  \multicolumn{1}{l}{---} &
  --- &
  Missing Certain word &
\AR{{تستضيف ← ان تستضيف }} 
   \\ \cmidrule{2-7}
 &
  XT &
  \multicolumn{1}{l}{Unnecessary Word} &
  \multicolumn{1}{l}{---} &
  --- &
  Adding Unnecessary Word &
  \AR{{حق الحالة ←  الحالة }} 
   \\ \cmidrule{2-7}

\multirow{-26}{*}{\rotatebox[origin=c]{90}{\textbf{\colorbox{red!10}{Syntax}}}}&
  XO &
  \multicolumn{1}{l}{Other Syntactic E.} &
  \multicolumn{1}{l}{---} &
  --- &
  Other Syntactic E. &
  \multicolumn{1}{r}{---} 
  \\ \cmidrule{1-7}
 
 &
   &
   &
   
  Removing Necessary Prep. &
  --- &
  C. in Removing Necessary Prep. &
  \AR{{ الربيع  ← للربيع}} 
   \\ \cmidrule{4-7}
 &
   &
   &
  Replacing Prep. &
  --- &
  C. in Replacing Prep. &
\AR{{  في سيارتنا ← بسيارتنا} } 
   \\ \cmidrule{4-7}
 &
   &
   &
  Adding Unnecessary Prep. &
  --- &
  C. in Adding Unnecessary Prep. &
\AR{{ من وراء ←  وراء }} 
   \\ \cmidrule{4-7}
 &
   &
   &
  Other Preposition E. &
  --- &
   &
  \multicolumn{1}{l}{} 
   \\ \cmidrule{4-7}
 &
   &
   &
  Preposition Use &
  --- &
  C. in Prep. Use &
 \AR{{ كطريقة ← طريقة }} 
   \\ \cmidrule{4-7}
 &
  \multirow{-6}{*}{SW} &
  \multirow{-6}{*}{Word Selection E.} &
  Other Word Selection E. &
  --- &
  C. in Other Word Selection E. &
  \multicolumn{1}{r}{---} 
   \\ \cmidrule{2-7}
 &
   &
   &
  Adding Unnecessary Conj. &
  --- &
  C. in Adding Unnecessary Conj. &
 \AR{{ حتى إلى ← إلى }} 
   \\ \cmidrule{4-7}
 &
   &
   &
 Removing Necessary Conj. &
  --- &
  C. in Removing Necessary Conjunction &
\AR{{المدير، الموظفون ← المدير والموظفون }} 
   \\ \cmidrule{4-7}
 &
   &
   &
  Replacing Conj. &
  --- &
  C. in Replacing Conj. &
 \AR{{فالمجتمع ← والمجتمع}} 
   \\ \cmidrule{4-7}
 &
  \multirow{-4}{*}{SF} &
  \multirow{-4}{*}{Conjunction E.} &
  Other Atf E. &
  --- &
  C. in Other Atf E. &
\AR{{مع مديره ←  ومديره}} 
   \\ \cmidrule{2-7}
 &
  SS &
  \multicolumn{1}{l}{Special Expression} &
  \multicolumn{1}{l}{---} &
  --- &
  C. in Special Expression &
  \AR{{هامة ← مهمة}} 
   \\ \cmidrule{2-7}
\multirow{-12}{*}{\rotatebox[origin=c]{90}{\textbf{\colorbox{orange!10}{Semantics}}}} &
  SO &
  \multicolumn{1}{l}{Other Semantic E.} &
  \multicolumn{1}{l}{---} &
  --- &
  Other Semantic E. &
  \multicolumn{1}{r}{---} 
  \\ \cmidrule{1-7}
 &
  MI &
  \multicolumn{1}{l}{Word Inflection} &
  \multicolumn{1}{l}{---} &
  --- &
  C. in Word Inflection &
 \AR{{مشينة ←  شـائنة}} 
   \\\cmidrule{2-7}
 &
  MT &
  \multicolumn{1}{l}{Verb Tense} &
  \multicolumn{1}{l}{---} &
  --- &
  C. in Verb Tense &
  \AR{{يعيش  ← عاش}} 
   \\\cmidrule{2-7}
\multirow{-4}{*}{\rotatebox[origin=c]{90}{\textbf{\colorbox{gray!10}{Morph}}}} &
  MO &
  \multicolumn{1}{l}{Other Morphological E.} &
  \multicolumn{1}{l}{---} &
  --- &
  Other Morphological E. &
  \multicolumn{1}{r}{---} 
  \\ \cmidrule{1-7}
 &
   &
   &
  Exclamation Mark (!) &
  --- &
   E. in Exclamation Marks. &
  \AR{{ما أجمل؟ ← ما أجمل!}} 
   \\
 &
   &
   &
  Semicolon (;) &
  --- &
  E. in Semicolon. &
  \AR{{ذهبت إلى الطبيب; ← ذهبت إلى الطبيب}} 
   \\
 &
   &
   &
  Colon (:) &
  --- &
  E. in Colon. &
 \AR{{ قال المعلم: ← قال المعلم: }} 
   \\
 &
   &
   &
  Ellipsis (…) &
  --- &
 E. in Ellipsis. &
\AR{{فيقول أحدهما.. ←  فيقول أحدهما...}} 
   \\ 
 &
   &
   &
  Quotations Marks (`` ”) &
  --- &
   E. in Quotation Marks. &
 \AR{{“ }} 
   \\ 
 &
   &
   &
  Square Brackets {[}{ ]} &
  --- &
 E. in Square Brackets. &
 \AR{ ← نعم}
{[}
\AR{نعم}
   \\
&
    &
    &
   Comma (,) &
   --- &
   E. in Comma. &
  \AR{{الشعر ،  ← الشعر}} 
   \\
  &
    &
    &
  Question Mark (?) &
  --- &
 E. in the Question Marks. &
\AR{{  كيف ذهبت. ← كيف ذهبت؟ }} 
   \\
 &
   &
   &
  Parenthesis () &
  --- &
  E. in Parenthesis. &
 \AR{{ مؤلفة (من ← مؤلفة من} } 
   \\
 &
   &
   &
  Dash (-) &
  --- &
   E. in the Dash. &
\AR{{  بعض -أفرادها ← بعض أفرادها }} 
   \\

\multirow{-11}{*}{\rotatebox[origin=c]{90}{\textbf{\colorbox{green!10}{Punctuation}}}}&
  \multirow{-11}{*}{\begin{tabular}[c]{@{}c@{}}PC\\ PM\\ PT\\ PO\end{tabular}} &
  \multirow{-11}{*}{\begin{tabular}[l]{@{}l@{}}Punctuation C.\\ Missing Punctuation\\ Unnecessary Punctuation\\ Other E.  in  Punctuation\end{tabular}} &
  Period/full stop (.) &
  --- &
  E. in full stop. &
 \AR{{يجب أن تقوم. انتفاضة ← يجب أن تقوم انتفاضة }} 
   \\ \cmidrule{1-7}

\colorbox{cyan!10}{\textbf{Merge}} &
  MG &
  \multicolumn{1}{l}{} &
  \multicolumn{1}{l}{---} &
  --- &
  Words are merged &
 \AR{{ إنشاء الله ← إن شاء الله }} 
   \\ \cmidrule{1-7}
\colorbox{purple!10}{\textbf{Split}} &
  SP &
  \multicolumn{1}{l}{} &
  \multicolumn{1}{l}{---} &
  --- &
  Words are split &
 \AR{{ لو لا ← لولا }} 
 \\
   \bottomrule
\end{tabular}
}
 \caption{The ALC error taxonomy has been expanded with new subcategories and subdivisions. Error tags are listed alphabetically, except for the `Other' tags, which are excluded in this extension. \textbf{The table includes abbreviations:} \textsuperscript{$\star$}\texttt{Chr.} for `Characters' \textsuperscript{$\star$}\texttt{Fem} for `Feminine' \textsuperscript{$\star$}\texttt{Masc.} for `Masculine' \textsuperscript{$\star$}\texttt{Sing.} for `Singular' \textsuperscript{$\star$}\texttt{Pl.} for `Plural' \textsuperscript{$\star$}\texttt{Num.} for `Number' \textsuperscript{$\star$}\texttt{Conj.} for `Conjunctions' \textsuperscript{$\star$}\texttt{Nom.} for `Nominative' \textsuperscript{$\star$}\texttt{Juss.} for `Jussive' \textsuperscript{$\star$}\texttt{Pron.} for `Pronouns' \textsuperscript{$\star$}\texttt{Acc.} for `Accusative' \textsuperscript{$\star$}\texttt{C} for `Confusion' and \textsuperscript{$\star$}\texttt{E} for 'Error(s).\textbf{\colorbox{gray!10}{Morph:}} Morphology, \texttt{Class:} Error Class, \texttt{Tag:} Error Tag, \texttt{Sub2:} Subclass of a subclass, \texttt{Sub3:} Subclass of a subclass of a subclass.}
 
 \label{tab:Extended Error Taxonomy}
\end{table*}

 \begin{table*}[h]
\centering
\resizebox{0.8\textwidth}{!}{%
\begin{tabular}{@{}m{0.35\linewidth}m{0.65\linewidth}@{}}
\toprule
\textbf{Morphological Template} &
  \textbf{Description}
   \\ \midrule
\textbf{Template 1:\AR{{فعل }} (fa’ala)} &
  Expresses the general verbal meaning of the root in question.
   \\ \hline
\textbf{Template 2: \AR{{فعّل}} (fa’’al)} &
  Built on template 1 by doubling the middle radical (adding a shadda to it). Often is an intensive or causative version of template 1.
   \\ \hline
\textbf{Template 3:\AR{{فاعل }} (faa’il)} &
  Built on template 1 by adding an alif between the first and second radicals. It gives an transitive or indicates a relation meaning to the form 1 verb, describes someone doing the act to or with someone else. 
   \\ \hline
\textbf{Template 4: \AR{{أفعل}} (af’al)} &
  Built on template 1 by prefixing an alif and putting a sukuun over the first radical. Similar to template 2 in that it is usually a causative or transitive version of template 1. 
   \\ \hline
\textbf{Template 5: \AR{{تفعّل}} (taff’al)} &
  Built on template 2 by adding the prefix \AR{{تـ}}(ta). Often a reflexive or passive version of template 2. 
   \\ \hline
\textbf{Template 6:\AR{{تفاعل }} (tafaa’al)} &
  Built on template 3 by adding the prefix \AR{{تـ}} (ta) . Usually a reflexive or passive version of template 3. 
   \\ \hline
\textbf{Template 7: \AR{{انفعل}} (infa’ala)} &
  Built on template 1 by adding the prefix \AR{{انـ}} (inna). Usually a reflexive and/or passive version of template 1. 
   \\ \hline
\textbf{Template 8:\AR{{افتعل }} (ifta’la)} &
  Built on template 1 by adding the prefix \AR{{ا}} (?a) and placing a sukuun must be placed over its first radical.It indicate a reflexive nuances or doing something intentionally version of the template 1. 
   \\ \hline
\textbf{Template 9:  \AR{{افعلّ}} (if’all)} & Built on template 1 by adding the prefix \AR{{ا}} (?a), placing a sukuun over its first radical, and adding a shadda to the last radical. 
   \\ \hline
\textbf{Template 10: \AR{{استفعل}} (istf’al)} &
  Built on template 1 by adding the prefix  \AR{{استـ}} (?sta) and inserting a\AR{ت}  (ta) between the first and second radicals, a sukuun must be placed over the first radical. Often a considerative version of template 1, means to consider or to deem someone in relation to template 1. 
   \\ \bottomrule
\end{tabular}
}
\caption{Synthetic error generation templates for word inflection based on the ATB.}
\label{tab:Synthetic_Data}
\end{table*}

 \textbf{Semantic Errors} refer to an error in the meaning or interpretation of a sentence or phrase, despite it being grammatically correct. This type of error occurs when words or phrases are used inappropriately, leading to confusion or a failure to convey the intended message. Semantic errors can stem from using a word that does not have the intended meaning of the sentence, or from a logical inconsistency in the statement. For instance, adding unnecessary conjunction as \AR{\small{قرأت الكتاب والذي كان مفيداً}}, the error here lies in the extra conjunction \AR{\small{واو}} in the sentence. Also, there is an error of missing the preposition \AR{\small{إلى}} in \AR{\small{محولة المنازل ثكنات عسكرية}}.  In  the semantics category, which initially comprised three subclasses, we added a new subclass designated \textit{Special Expressions}. This subclass addresses the incorrect usage of special expressions at either the word or phrase level. Breakdown of this category includes a Word Selection Error subclass divided into five further categories: i) \textit{Adding unnecessary preposition}, ii) \textit{Removing necessary preposition}, iii) \textit{Replacing preposition}, iv) \textit{Preposition use}, v) and \textit{other word selection errors}. Additionally, for conjunction errors, we organized this into four sub-subclasses: i) \textit{Adding unnecessary conjunction}, ii) \textit{Removing necessary conjunction}, iii) \textit{Replacing conjunction}, iv) and \textit{other Atf errors}. The subsubclasses we introduced in the semantic errors might appear to overlap with the missing and additional characters in the syntax class, but the errors we generated impact the sentence's semantic meaning rather than its structural integrity.\\

 \textbf{Punctuation Errors} occur when punctuation marks are misused, omitted, or placed incorrectly within sentences. These errors can alter the clarity, flow, and meaning of text. Typical punctuation errors include the misuse of commas, periods, semicolons, colons, and quotation marks, etc. For instance, the confusion of using question mark in \AR{\small{ما أروع القمر؟}} instead of using the exclamation mark \AR{\small{ما أروع القمر!}}. Within the framework of the original error taxonomy, punctuation errors are categorized into four subclasses: i) \textit{missing punctuation}, ii) \textit{punctuation confusion}, iii) \textit{unnecessary punctuation}, iv) and \textit{other punctuation errors}. These subclasses are further divided into detailed sub subclasses that specifically address various punctuation marks, such as commas, exclamation marks, and full stops, etc.\\
 
\section{ATB Synthetic Data}
\label{apdx:synthetic_data}
 \textbf{Orthographic Errors}
\begin{itemize}
     \item \textbf{Missing Characters (OM)} In generating OM errors, we apply three distinct types of rules. The initial approach involves randomly removing a character, with each character having an equal probability of being removed, or following a distribution where certain characters like \AR{\small{م}}, \AR{\small{ي}}, \AR{\small{ئ}}, \AR{\small{ل}} are more likely to be removed than others. 
     The second rule focuses on removing a character in instances of repetition.
      For example: \AR{\small{لعملية}} would be \AR{\small{لعلية}} and \AR{\small{جييا}} would be \AR{\small{جيا}} 
  
  The third rule involves removing words that are written but not pronounced, such as: \AR{\small{الألف الشمسية}},  Hamzet al wasel \AR{\small{همزة الوصل}} after \AR{\small{و}} and \AR{\small{ف}}, Alif Tanwin \AR{\small{ألف التنوين}},
  \AR{\small{و}} after Hamza \AR{\small{ئ}} and \AR{\small{ر}}.

    \item \textbf{Wrong Order of Word Characters (OC)} 
    When generating OC errors, we either shuffle two adjacent characters within the same word or find a word that shares the same characters as the target word aimed to change and has a Levenshtein distance of 2. For example \AR{\small{يجب}} would be \AR{\small{جيب}}
    
  \item \textbf{Replacement in word characters (OR)}
  In generating OR errors, we pursue two distinct methods.
  The first one, we seek a word with a Levenshtein distance of 1 from the target word. For the second one, we randomly select a character and replace it with another character based on specific criteria: 1- Characters that are visually similar in shape, 2- Characters that are located near each other on the keyboard, 3- Characters that have similar sounds or are phonetically close when pronounced.
   For example, if we change the character \AR{\small{ا}}, we might select \AR{\small{آ}}, \AR{\small{إ}}, or \AR{\small{أ}} as alternatives. For the character \AR{\small{ط}}, possible substitutions include \AR{\small{ت}}, \AR{\small{ظ}}, \AR{\small{ص}}, \AR{\small{ض}}, \AR{\small{ك}}, or \AR{\small{د}}. 
    For instance, the word \AR{\small{البرتغال}} could be altered to \AR{\small{البرتقال}}. The character \AR{\small{ق}} is replaced with \AR{\small{غ}}, as they are similar in shape and located close to each other on the keyboard.
  
  \item \textbf{Missing Characters (OM))}
  In generating OM errors, we follow three distinct techniques. The first technique involves adding characters that are pronounced but should not be written. For instance, in the case of Alif al mad \AR{\small{ألف المد}} and Tanwin \AR{\small{التنوين}}, the Alif is pronounced in the word \AR{\small{هذا}}, but it should not be written. Thus, the resulting error word is \AR{\small{هاذا}}. The second one involves randomly adding characters. For example, the word \AR{\small{ففي}} could be altered to \AR{\small{في}}. The third technique requires searching for a word that has a Levenshtein distance of 1 from the target word and is shorter in length. For example, \AR{\small{معه}} could be changed to \AR{\small{مع}}. \\

\textbf{Syntax Errors}
\item \textbf{Missing Word (XM)} In generating XM errors, we employ the following strategies:
1- remove pronouns such as such as demonstrative, reflexive pronouns, ...etc. 2- remove conjunctions, prepositions, or inna \AR{\small{ان}} 3- remove bi-gram collocation words. For instance \AR{\small{في الدور}} would be altered to \AR{\small{الدور}}.

\item \textbf{Unnecessary Word (XT)} In XT error generation, we employ the following strategies:
1- Add conjunctions and prepositions. 2- Include unnecessary pronouns. For example, in the phrase \AR{\small{نلعب نحن}}, the pronoun \AR{\small{نحن}} is redundant since it is already indicated by the \AR{\small{ن}} in the verb \AR{\small{لعب}}. 3- Add bi-gram collocation words. For instance, \AR{\small{ما يطبع}} could be modified to \AR{\small{ما يحدث يطبع}}.
\end{itemize}  
 
\textbf{Morphological Errors}
\begin{itemize}
\item In Morphological errors, there are two types of error generated: word inflection and verb tense. 
Starting with word inflection, in the Arabic language most of the words are derived from three literal root which represents a core meaning or concept. There are cases for four, and five literal root, but the most common one is three literal root. By using the root, a variety of nouns, verbs, adjectives, active participles, and passive participles could be derived by adding specific prefixes, suffixes and using special templates. Each template has a basic meaning associated with the general meaning of the root being used.
To derive word inflection errors, we transform any noun, verb, or adjective to another one by changing its template while preserving its meaning by checking the gloss of the original word and generated inflection.
We have used the certain templates to derive inflection on words.For example, An example of inflection \AR{\small{الشهيرة}} would be \AR{\small{المشهورة}}. Table \ref{tab:Synthetic_Data} shows the employed templates.

\item In Arabic verb tense errors, verbs are modified using specific suffixes and prefixes to indicate gender, number, and person. To transform one verb tense to another, these suffixes and prefixes must be adjusted accordingly.
For example, to change the past verb \AR{\small{وضعنا}} to its present form, the suffix \AR{\small{نا}} is removed, and the prefix \AR{\small{ن}} is added.
Another example is transforming the past verb \AR{\small{التقتا}} to its present form \AR{\small{تلتقيان}}. Here, the suffix \AR{\small{تا}} (indicating a feminine dual third person) is removed, and the suffix \AR{\small{ان}} (indicating dual form) is added, along with the prefix \AR{\small{ت}} to indicate a feminine third person.
To convert a verb to future tense, it should first be transformed into the present tense, after which the prefix \AR{\small{س}} is added.
\end{itemize}
\textbf{Merge Error}     
In generating merge errors, we consider two main approaches:
1- Randomly merging two consecutive words together. For example, \AR{\small{فما كان}} would be 
\AR{\small{فماكان}}.
2- Merging specific types of words, which includes:
a) Merging prepositions with the following word.  For example, \AR{\small{عن هذا}} would be \AR{\small{عنهذا}}.
b) Merging \AR{\small{إن}} with the next word.
c) Merging negation words with the next word.
d) Merging vocative cases with the next word.
e) Merging nominal or personal pronouns with the next word. For example \AR{\small{هذه الدول}} would be \AR{\small{هذهالدول}}. \\

\textbf{Split Error}
In generating split errors, we consider two main approaches:
1- Randomly splitting any word at any position. For example, \AR{\small{اهدافه}} could be split into \AR{\small{اهدا فه}}.
2- Splitting specific types of words, such as:
a) Interrogative words: \AR{\small{فيما}} would become \AR{\small{في ما}}.
b) Conjunctions: \AR{\small{ونفت}} would be split into \AR{\small{و نفت}}.
c) Pronouns: For instance, \AR{\small{الفلاحين}} would be split into \AR{\small{الفلاح ين}}.\\
\textbf{Punctuation Error}
In generating punctuation errors, we consider the following categories:
1- Missing punctuation: We randomly remove punctuation marks from sentences.
2- Unnecessary punctuation: We add random punctuation marks at any position within the sentence.
3- Punctuation confusion: We replace punctuation based on shape and keyboard placement. For example:
The comma ``،" is most likely to be replaced with ``," or ``;" or ``.". The closing bracket``]" is likely to be replaced with ``[", ``(" or ``\}".
\label{atb_synthetic_data_appdx}

\section{Prompt Design}
\label{apdx:prompt_design}
\begin{table*}[b]
\resizebox{\textwidth}{!}{%
\begin{tabular}{cllclcrr}
\toprule
\multicolumn{7}{c}{\textbf{Gazelle Dataset}}                                                                                                                                                                 \\ \midrule
\multicolumn{2}{c}{\colorbox{gray!10}{\textbf{ALC Corpus}}}                    &          & \multicolumn{2}{c}{\colorbox{green!10}{\textbf{New categories}}} & & \multicolumn{2}{c}{\textbf{Examples}}                                                                                                                                                                                                  \\ \cmidrule{1-2} \cmidrule{4-5} \cmidrule{7-8}
\colorbox{gray!10}{\textbf{Type} }                 & \colorbox{gray!10}{\textbf{Tag}}       &                    & \colorbox{green!10}{\textbf{Subclass}}                                 & \multicolumn{1}{c}{\colorbox{green!10}{\textbf{Sub2}}}       &  & \multicolumn{1}{c}{\textbf{Text with errors}}                                    & \multicolumn{1}{c}{\textbf{Corrected text}}                                 \\ \midrule

                                        &                                              &                 & & Medial Hamza                    & &
                                        \n \AR{هامة.}
                                        \r \AR{ مسالة} 
                                       \n \AR{كانت تلك} 
                                        &    \n \AR{هامة.}
                                        \g \AR{ مسألة} 
                                         \n\AR{كانت تلك}                                             \\
                                        &                                              &                                                      &   & Hamzet Wasel                &    &      \n \AR{ بالمطر خيراً.} \r \AR{إستبشرت}              & \n \AR{ بالمطر خيراً.} \g \AR{استبشرت}                           \\
                                        &                                              &                                                     &    & The Cutting Hamzah             & & \n \AR{السكنات} \r\AR{احد} \n \AR{بخصوص تأمين سكن وظيفي في}              & \n \AR{السكنات} \g \AR{أحد} \n \AR{بخصوص تأمين سكن وظيفي في}          \\
\multirow{-4}{*}{\rotatebox[origin=c]{90}{\textbf{Orth.}}} & \multirow{-3}{*}{OH}                       &  & \multirow{-4}{*}{Hamza}                                 & Final Hamza    &                 & \n \AR{ المسلم على الكذب.} \r \AR{يجروء}    \n\AR{لا}           & \n\AR{ المسلم على الكذب.} \g\AR{يجرؤ}   \n\AR{لا}                     \\ \midrule
                                        &                                              &                           &                              & Masculine Sound Plural       &   & \n\AR{ السيارات ملتزمون بتعليمات إدارة المرور.} \r\AR{سائقوا}& \n\AR{ السيارات ملتزمون بتعليمات إدارة المرور.} \g\AR{سائقو} \\ 
                                        &                      &                        & \multirow{-2}{*}{Plural}                                & Almaqsour Plural              &  & \n\AR{ لا ندركه.}\r\AR{ بُلَهَاء}\n\AR{وعلى مرمى طوبة زمن منا ونحن لا نعيه،} &  \n\AR{ لا ندركه.}\g\AR{ بُلْهٌ}\n\AR{وعلى مرمى طوبة زمن منا ونحن لا نعيه،} \\\cmidrule{4-8}
                                        &               &                               &                                                         & Composite Numbers            &  &\r\AR{ عشر}\n\AR{يمكنكم الآن مشاهدة الحلقة الثالثة} & \g\AR{ عشرة}\n\AR{يمكنكم الآن مشاهدة الحلقة الثالثة}  \\
\multirow{-4}{*}{\rotatebox[origin=c]{90}{\textbf{Syntax}}}       & \multirow{-4}{*}{XN}            &             & \multirow{-2}{*}{Number}                                & Al Aqoud                   &  & \n\AR{ من عمره.}\r\AR{ الخمسون}\n\AR{حضر مريض في}&  \n\AR{ من عمره.}\g\AR{ الخمسين}\n\AR{حضر مريض في} \\ \midrule
       & \multirow{4}{*}{SW}            &             & \multirow{4}{*}{Word Selection Error}                                & Adding Unnecessary Prepositions                   &  & \n\AR{ الأمر.} \r\AR{عن} \n\AR{تحرى فلان}  & \n\AR{تحرى فلان الأمر.}\\ \cmidrule{5-8}

&               &                               &                                                         & Removing Necessary Prepositions             &  & \n\AR{تسعى الدولة تحقيق مكاسب دولية.}  & \n\AR{ تحقيق مكاسب دولية.}\g\AR{إلى} \n\AR{تسعى الدولة} \\ \cmidrule{5-8}

&               &                               &                                                         & Replacing Prepositions             &  &  \n\AR{ بلاد مختلفة.}\r\AR{في} \n\AR{ثم وجدت طلبا}& \n\AR{بلاد مختلفة.} \g\AR{من} \n\AR{ثم وجدت طلبا} \\ \cmidrule{2-8}
\multirow{-5}{*}{\rotatebox[origin=c]{90}{\textbf{Semantics}}}       & \multirow{3}{*}{SF}            &             & \multirow{3}{*}{Conjuction Error}                                & Adding Unnecessary Conjunction                   &  &\n\AR{ إلى قلوب بعض المسلمين والمسلمات.} \r\AR{حتى}\n\AR{ووصلت صدى هذه الأصوات } & \n\AR{  قلوب بعض المسلمين والمسلمات.} \g\AR{إلى} \n\AR{ووصلت صدى هذه الأصوات} \\ \cmidrule{5-8}
                                            &               &                               &                                                         & Removing Necessary Conjunction             &  &\n\AR{له كذا.} \r\AR{وقل} \n\AR{اذهب إلى فلان} & \n\AR{له كذا.} \g\AR{فقل} \n\AR{اذهب إلى فلان}\\ \cmidrule{1-8}

\multirow{4}{*}{\rotatebox[origin=c]{90}{\textbf{Punc.}}}       & \multirow{1}{*}{PO}            &             & Other Punctuation                               &      Comma (,)      &  &\n\AR{عبّر عن رأيه بأسلوب سلس.} \r\AR{ ، } \n\AR{فضلًا عن ذلك } & \n\AR{عبّر عن رأيه بأسلوب سلس.} \g\AR{،} \n\AR{فضلًا عن ذلك} \\ \cmidrule{2-8}
                                            & \multirow{1}{*}{PC}            &             & Commission Errors                               &          Full Stop (.)  &  & \r\AR{؟} \n\AR{الشمس تزود الأرض بالحرارة والضوء}& \n\AR{الشمس تزود الأرض بالحرارة والضوء.} \\ \cmidrule{2-8}
                                            & \multirow{1}{*}{PM}            &             & Misuse of Punctuation Marks                               &        Question Marks (?)    &  &\r\AR{رغبته} \n\AR{هل يستطيع هذا القس أن يعلن } & \g\AR{؟} \n\AR{هل يستطيع هذا القس أن يعلن رغبته} \\

\bottomrule
\end{tabular}%
}
\caption{Comparison of \textit{Gazelle} with the QALB 2014, QALB 2015, and ZAEBUC datasets, all of which are based on the ALC error taxonomy. \textit{Gazelle} provides more fine-grained and specific error type coverage within this taxonomy.\textsuperscript{$\star$}\texttt{Sub2} represents our extended subclass within the ALC error types. \textsuperscript{$\star$}\texttt{Orth.:} Orthography \textsuperscript{$\star$}\texttt{Punc.:} Punctuation.}
\label{comparsion}
\end{table*}

Designing effective prompts is fundamental to leveraging the capabilities of LLMs. A prompt, defined as a set of instructions given to an LLM, acts as a programming tool that tailors the model’s responses to specific tasks~\cite{white2023prompt}. By establishing a set of specific rules and intents, prompts can significantly influence both the interactions with the model and its generated outputs. Thus, to achieve the desired outcomes, it is essential to carefully construct prompts that clearly define the task's objectives and guidelines. 

To create effective prompts for our writing tasks, we designed a set of zero-shot prompts tailored to each subtask to elicit the desired responses from the models. For instance, in the I’rab task, each model was instructed to provide the complete I’rab of the sentences and explain their case markers. Similarly, for the MWEs task, models were prompted to identify and explain MWEs within sentences, including their meanings and contextual usage.

We sample 2 sentences for each subtask and use the web interfaces of all the models for the initial evaluation, focusing on the effectiveness of the prompts in eliciting the desired responses. By assessing and comparing the outputs generated by these initial prompts, we identify the most effective prompt. For most subtasks, the prompts we design successfully meet our requirements for the generated outputs.

\begin{figure*}
    \centering
    \includegraphics[width=1\linewidth]{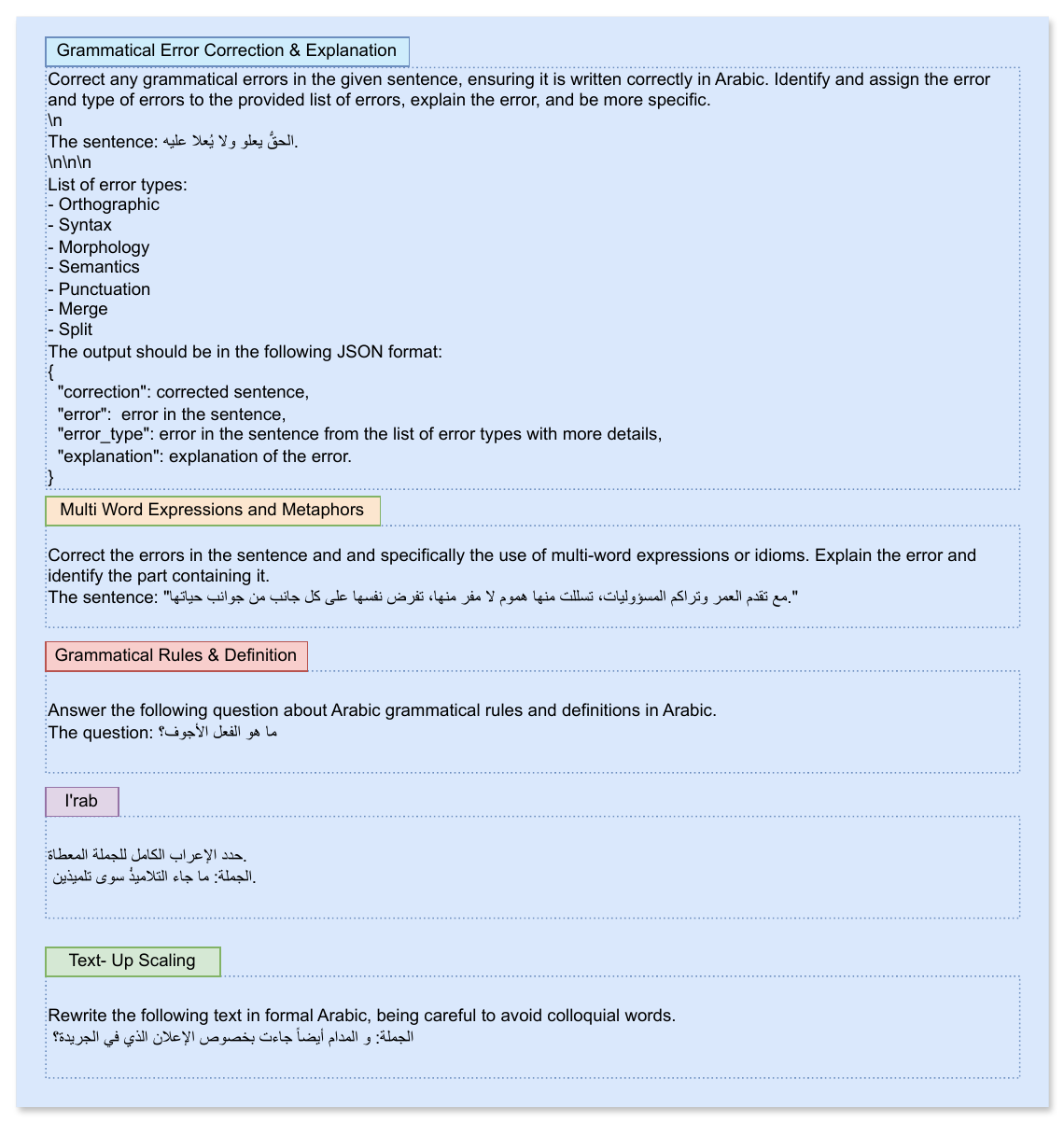}
    \caption{Prompts used  for LLMs evaluation.}
    \label{fig:prompt_template}
\end{figure*}

However, the GEC prompt required a more detailed approach. Initially, we developed a zero-shot prompt instructing the models to identify the type of grammatical error in a given sentence without specifying the error types, using the prompt: "I will give you an MSA sentence with a grammatical error. You should identify the type of error, provide the correct version of the sentence, and explain the error in Arabic. The sentence:". We then use GPT-4o to iteratively refine this prompt, enhancing its ability to identify the type of error from a specified list, provide the correct version of the sentence, and explain the error. After refining, we modify the prompt and standardize it to fit all models. Detailed prompt templates we use can be found in Table~\ref{fig:prompt_template}.

\section{Evaluation matrices for Arabic Writing Assistant Tasks}
As mentioned earlier how we come up with the evaluation metrics of six tasks of Arabic writing assistance tasks, we provide detailed description of the metrics we employ in our evaluation. 
For the GEC and explanation tasks for example, we assess three specific criteria: accuracy, clarity, and helpfulness.

\noindent \textbf{Accuracy} refers to "the degree to which the response correctly addresses the question or task, without errors in information or execution." This criterion is rated on a scale from 1 to 5 as follows:
 \begin{itemize}
     \item 5: Completely accurate with no errors.
     \item 4: Mostly accurate with minor errors.
     \item 3: Moderately accurate with some errors.
     \item 2: Somewhat accurate with several errors.
     \item 1: Inaccurate with many errors.
 \end{itemize}
 
\noindent\textbf{Clarity} indicates the ease with which the response can be understood, including how well it conveys the intended message without ambiguity. This criterion is rated on the following scale:
 \begin{itemize}
     \item 5: Extremely clear and easy to understand.
     \item 4: Mostly clear with minor ambiguities.
     \item 3: Moderately clear with some ambiguities.
     \item 2: Somewhat unclear with several ambiguities.
     \item 1: Very unclear and difficult to understand.
 \end{itemize}
 
\noindent\textbf{Helpfulness} measures "the extent to which the response aids the user in understanding or resolving the issue, providing useful and actionable information." The scale for this criterion is as follows:
 \begin{itemize}
     \item 5: Extremely helpful and informative.
     \item 4: Mostly helpful with minor gaps in information.
     \item 3: Moderately helpful with some gaps in information.
     \item 2: Somewhat helpful with several gaps in information.
     \item 1: Not helpful and lacking useful information.
 \end{itemize}

\begin{figure*}[t]
    \centering
    \begin{subfigure}[b]{15cm}
        \centering
        \includegraphics[width=15cm,height = 4cm]{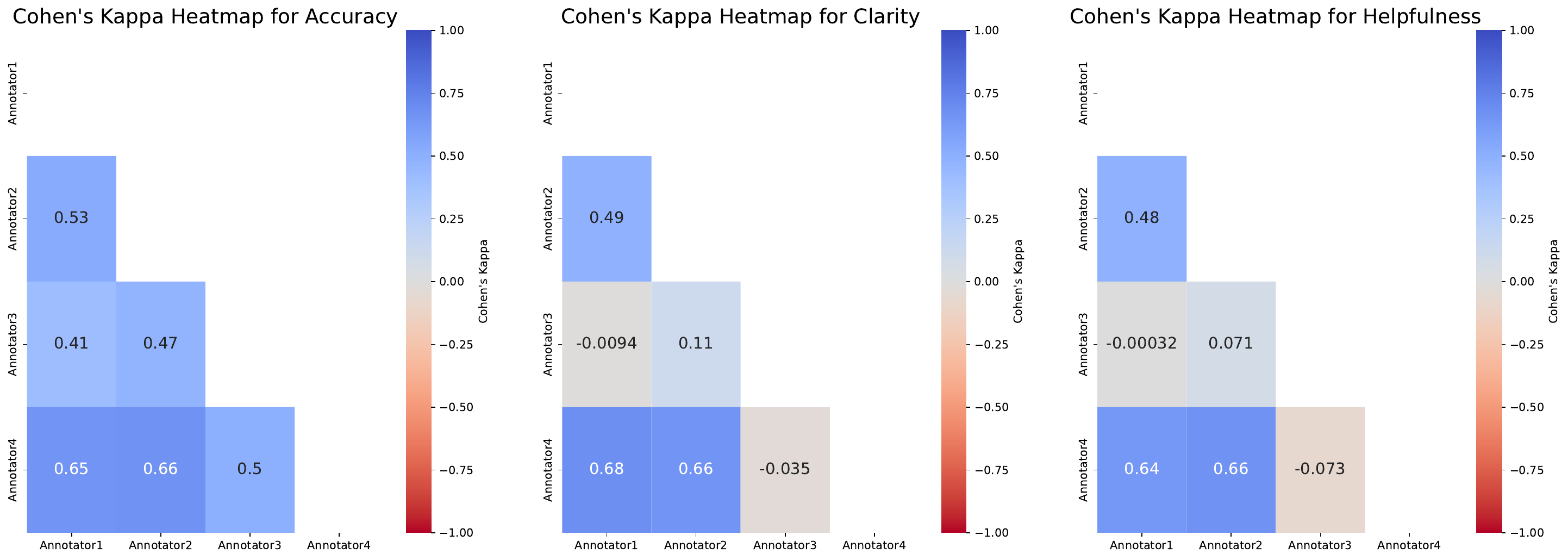}
        \caption{GEC+ Explanation task inter-annotators agreement.}
        \label{fig:gec_agreement}
    \end{subfigure}
    \vfill
    \begin{subfigure}[b]{15cm}
        \centering
        \includegraphics[width=15cm,height = 4cm]{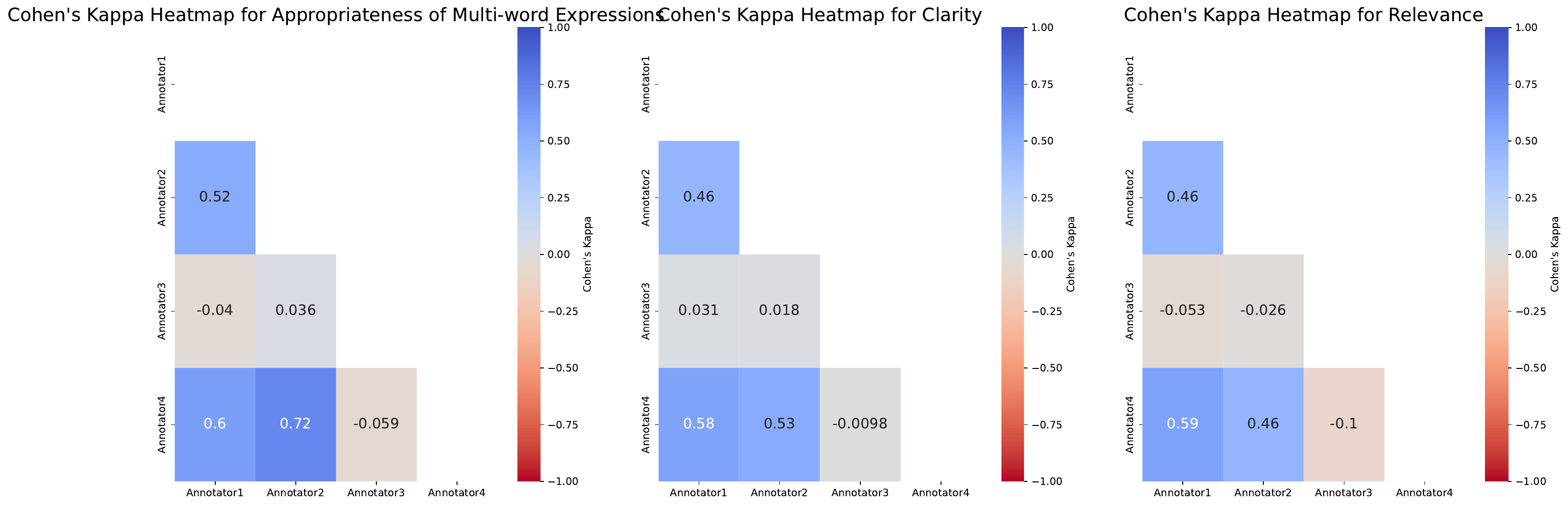}
        \caption{MWE task inter-annotators agreement.}
        \label{fig:mwe_agreement}
    \end{subfigure}
    \vfill
    \begin{subfigure}[b]{15cm}
        \centering
        \includegraphics[width=15cm,height = 4cm]{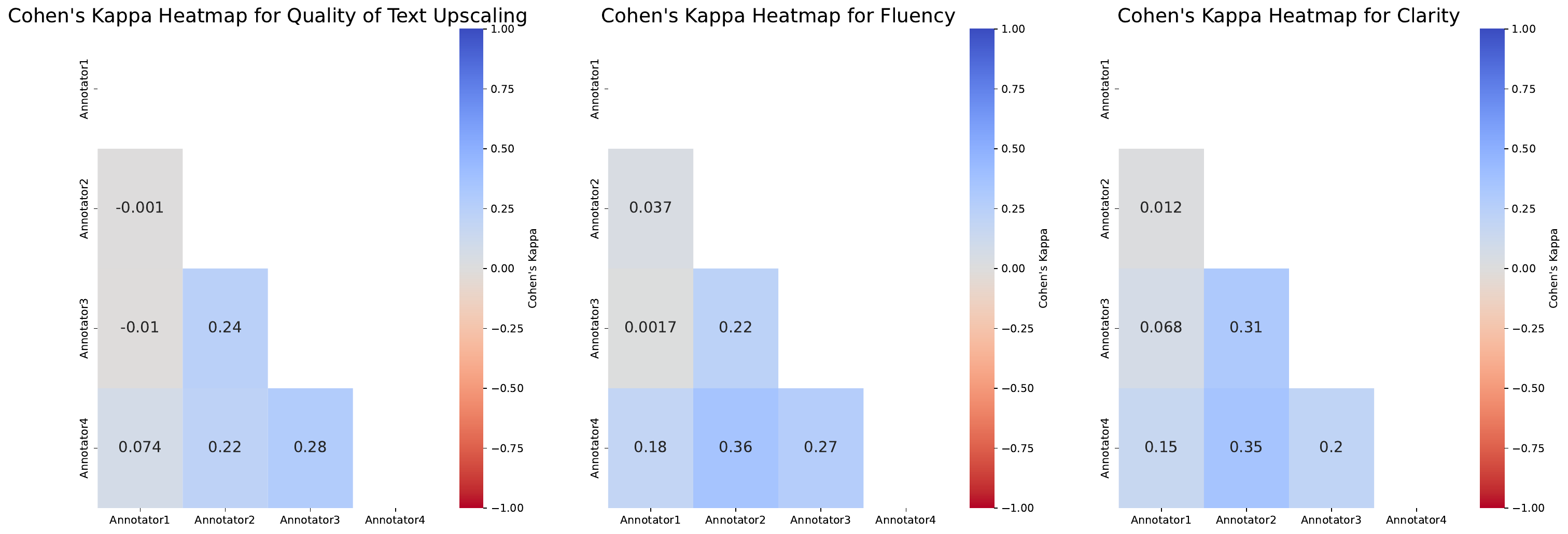}
        \caption{Text refinement task inter-annotators agreement.}
        \label{fig:text_agreement}
    \end{subfigure}
    \begin{subfigure}[b]{15cm}
        \centering
        \includegraphics[width=15cm,height = 4cm]{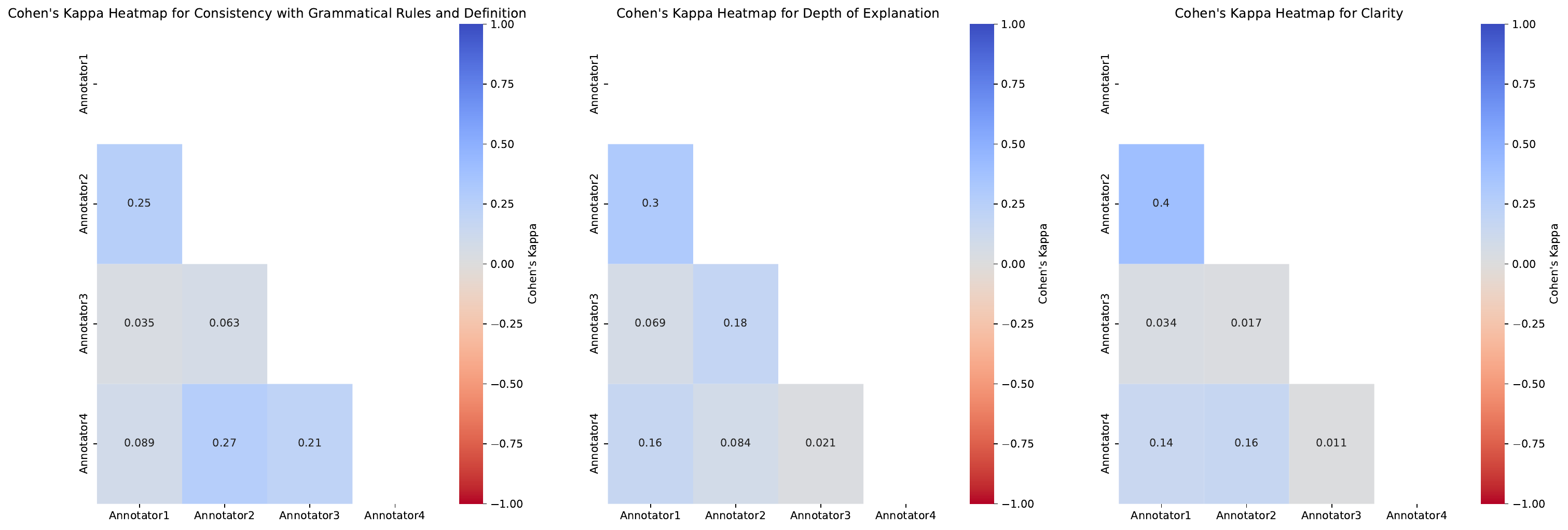}
        \caption{Grammatical Rules task inter-annotators agreement.}
        \label{fig:rules_agreement}
    \end{subfigure}
    \vfill
    \begin{subfigure}[b]{15cm}
        \centering
        \includegraphics[width=15cm,height = 4cm]{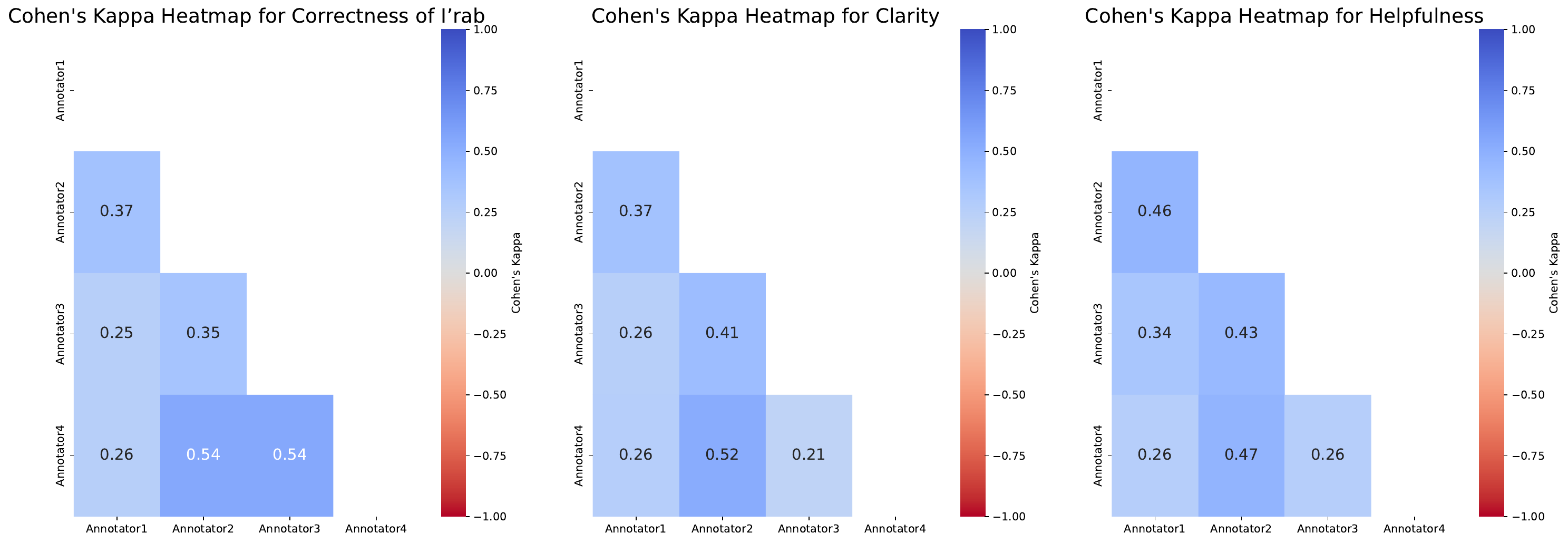}
        \caption{I'rab task inter-annotators agreement.}
        \label{fig:iraab_agreement}
    \end{subfigure}

    \caption{ Inter-annotator agreement for human evaluation using Cohen’s Kappa.}
    \label{fig:agreement}
\end{figure*}

\begin{landscape}
\begin{figure*}[p]
    \centering
    \includegraphics[width=\linewidth, height=\textheight, keepaspectratio]{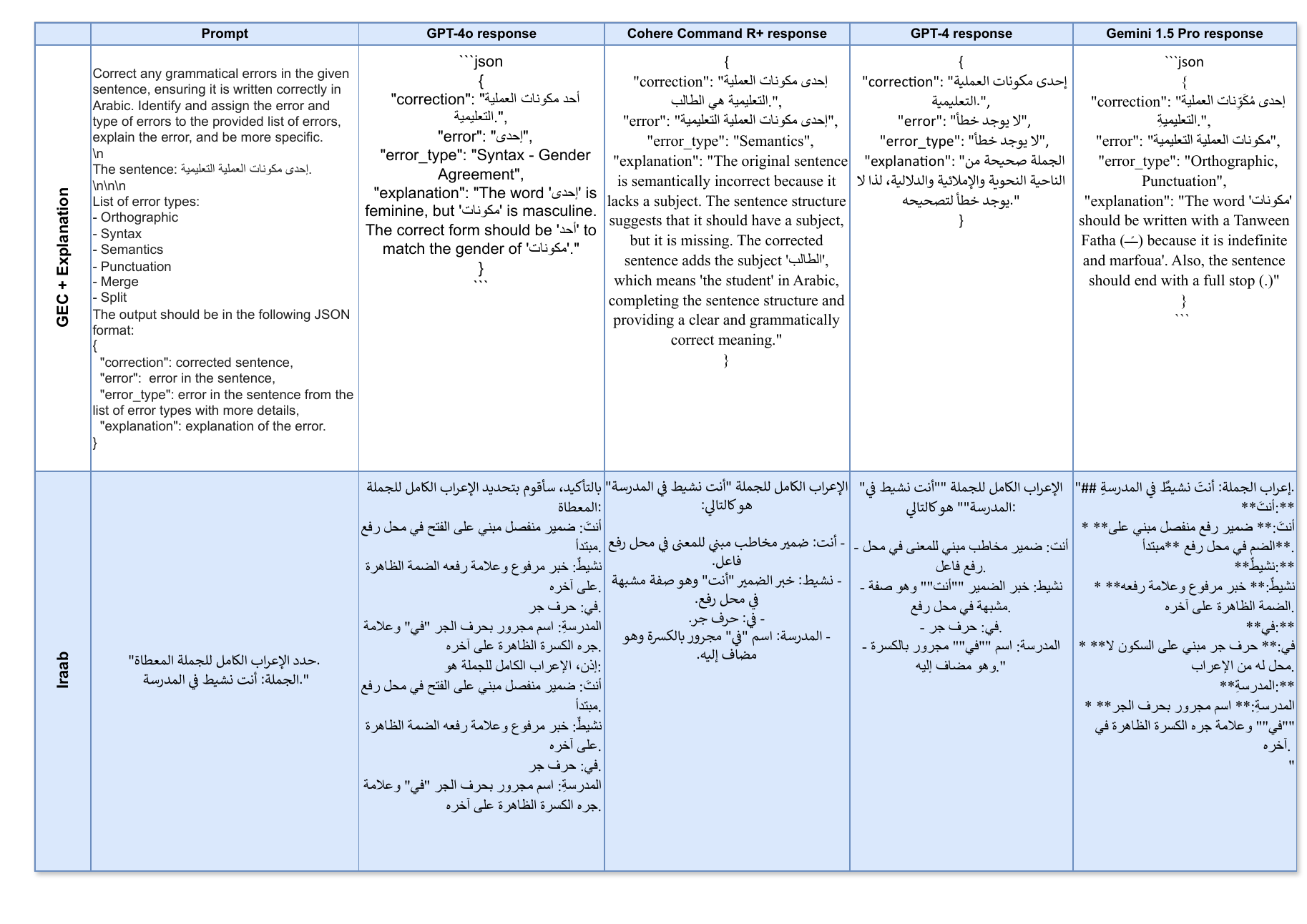}
    \caption{Sample model responses for GEC, explanation, and I'rab tasks.}
    \label{fig:models_responses}
\end{figure*}
\end{landscape}

The rest of the metrics are listed in Table \ref{tab:matrices}. Figure \ref{fig:agreement} illustrates the inter-annotator agreement for human evaluation using Cohen’s Kappa metrics. Figure \ref{fig:models_responses} shows examples of responses from the four models across two different tasks.

\label{apdx:matrices}

\begin{table*}
    \centering

\scalebox{0.65}{

\begin{tabular}{|>{\raggedright\arraybackslash}m{1cm}|p{4cm}|p{12cm}|}
\hline
\textbf{Task} & \textbf{Measures} & \textbf{Scale Details} \\ \hline
\parbox[t]{2mm}{\multirow{9}{*}{\rotatebox[origin=c]{90}{Grammar Error Correction+Explanation }}}
 & \multirow{5}{4cm}{Accuracy} & \cellcolor{green!25} 5: Completely accurate with no errors. \\ 
 &  & \cellcolor{yellow!25} 4: Mostly accurate with minor errors. \\ 
 &  & \cellcolor{yellow!50} 3: Moderately accurate with some errors. \\ 
 &  & \cellcolor{orange!50} 2: Somewhat accurate with several errors. \\ 
 &  & \cellcolor{red!50} 1: Inaccurate with many errors. \\ \cline{2-3} 
 & \multirow{5}{4cm}{Clarity} & \cellcolor{green!25} 5: Extremely clear and easy to understand. \\ 
 &  & \cellcolor{yellow!25} 4: Mostly clear with minor ambiguities. \\ 
 &  & \cellcolor{yellow!50} 3: Moderately clear with some ambiguities. \\ 
 &  & \cellcolor{orange!50} 2: Somewhat unclear with several ambiguities. \\ 
 &  & \cellcolor{red!50} 1: Very unclear and difficult to understand. \\ \cline{2-3} 
 & \multirow{5}{4cm}{Helpfulness} & \cellcolor{green!25} 5: Extremely helpful and informative. \\ 
 &  & \cellcolor{yellow!25} 4: Mostly helpful with minor gaps in information. \\ 
 &  & \cellcolor{yellow!50} 3: Moderately helpful with some gaps in information. \\ 
 &  & \cellcolor{orange!50} 2: Somewhat helpful with several gaps in information. \\ 
 &  & \cellcolor{red!50} 1: Not helpful and lacking useful information. \\ \hline
\parbox[t]{2mm}{\multirow{15}{*}{\rotatebox[origin=c]{90}{Multi-word Expressions }}}
& \multirow{5}{4cm}{Appropriateness of Multi-word Expressions} & \cellcolor{green!25} 5: Extremely appropriate and contextually fitting. \\ 
 &  & \cellcolor{yellow!25} 4: Mostly appropriate with minor issues. \\ 
 &  & \cellcolor{yellow!50} 3: Moderately appropriate with some issues. \\ 
 &  & \cellcolor{orange!50} 2: Somewhat appropriate with several issues. \\ 
 &  & \cellcolor{red!50} 1: Inappropriate and contextually unfitting. \\ \cline{2-3} 
 & \multirow{5}{4cm}{Clarity} & \cellcolor{green!25} 5: Extremely clear and easy to understand. \\ 
 &  & \cellcolor{yellow!25} 4: Mostly clear with minor ambiguities. \\ 
 &  & \cellcolor{yellow!50} 3: Moderately clear with some ambiguities. \\ 
 &  & \cellcolor{orange!50} 2: Somewhat unclear with several ambiguities. \\ 
 &  & \cellcolor{red!50} 1: Very unclear and difficult to understand. \\ \cline{2-3} 
 & \multirow{5}{4cm}{Relevance} & \cellcolor{green!25} 5: Completely relevant to the task/question. \\ 
 &  & \cellcolor{yellow!25} 4: Mostly relevant with minor off-topic elements. \\ 
 &  & \cellcolor{yellow!50} 3: Moderately relevant with some off-topic elements. \\ 
 &  & \cellcolor{orange!50} 2: Somewhat relevant with several off-topic elements. \\ 
 &  & \cellcolor{red!50} 1: Irrelevant to the task/question. \\ \hline
 \parbox[t]{2mm}{\multirow{17}{*}{\rotatebox[origin=c]{90}{I’rab }}}

& \multirow{5}{4cm}{Correctness of I'rab}& \cellcolor{green!25} 5: Completely correct with no errors.\\ 
 &  & \cellcolor{yellow!25} 4: Mostly correct with minor errors. \\ 
 &  & \cellcolor{yellow!50} 3: Moderately correct with some errors. \\ 
 &  & \cellcolor{orange!50} 2: Somewhat correct with several errors. \\ 
 &  & \cellcolor{red!50} 1: Incorrect with many errors. \\ \cline{2-3} 
 & \multirow{5}{4cm}{Clarity} & \cellcolor{green!25} 5: Extremely clear and easy to understand. \\ 
 &  & \cellcolor{yellow!25} 4: Mostly clear with minor ambiguities. \\ 
 &  & \cellcolor{yellow!50} 3: Moderately clear with some ambiguities. \\ 
 &  & \cellcolor{orange!50} 2: Somewhat unclear with several ambiguities. \\ 
 &  & \cellcolor{red!50} 1: Very unclear and difficult to understand. \\ \cline{2-3} 
 & \multirow{5}{4cm}{Helpfulness} & \cellcolor{green!25} 5: Extremely helpful and informative. \\ 
 &  & \cellcolor{yellow!25} 4: Mostly helpful with minor gaps in information. \\ 
 &  & \cellcolor{yellow!50} 3: Moderately helpful with some gaps in information. \\ 
 &  & \cellcolor{orange!50} 2: Somewhat helpful with several gaps in information. \\ 
 &  & \cellcolor{red!50} 1: Not helpful and lacking useful information. \\ \hline
  \parbox[t]{2mm}{\multirow{15}{*}{\rotatebox[origin=c]{90}{Grammatical Rule + Definitions }}}
& \multirow{5}{4cm}{Consistency with Grammatical Rules and Definitions} & \cellcolor{green!25} 5: Completely consistent and correct. \\ 
 &  & \cellcolor{yellow!25} 4: Mostly consistent with minor inconsistencies. \\ 
 &  & \cellcolor{yellow!50} 3: Moderately consistent with some inconsistencies. \\ 
 &  & \cellcolor{orange!50} 2: Somewhat consistent with several inconsistencies. \\ 
 &  & \cellcolor{red!50} 1: Inconsistent and incorrect. \\ \cline{2-3} 
 & \multirow{5}{4cm}{Depth of Explanation} & \cellcolor{green!25} 5: Extremely thorough and detailed. \\ 
 &  & \cellcolor{yellow!25} 4: Mostly thorough with minor details missing. \\ 
 &  & \cellcolor{yellow!50} 3: Moderately thorough with some details missing. \\ 
 &  & \cellcolor{orange!50} 2: Somewhat thorough with several details missing. \\ 
 &  & \cellcolor{red!50} 1: Superficial and lacking in detail. \\ \cline{2-3} 
 & \multirow{5}{4cm}{Clarity} & \cellcolor{green!25} 5: Extremely clear and easy to understand. \\ 
 &  & \cellcolor{yellow!25} 4: Mostly clear with minor ambiguities. \\ 
 &  & \cellcolor{yellow!50} 3: Moderately clear with some ambiguities. \\ 
 &  & \cellcolor{orange!50} 2: Somewhat unclear with several ambiguities. \\ 
 &  & \cellcolor{red!50} 1: Very unclear and difficult to understand. \\ \hline
 \parbox[t]{2mm}{\multirow{15}{*}{\rotatebox[origin=c]{90}{Text Upscaling }}}
& \multirow{5}{4cm}{Quality of Text Upscaling} & \cellcolor{green!25} 5: Extremely high quality with significant improvement. \\ 
 &  & \cellcolor{yellow!25} 4: High quality with notable improvement. \\ 
 &  & \cellcolor{yellow!50} 3: Moderate quality with some improvement. \\ 
 &  & \cellcolor{orange!50} 2: Low quality with minimal improvement. \\ 
 &  & \cellcolor{red!50} 1: Poor quality with no improvement or degradation. \\ \cline{2-3} 
 & \multirow{5}{4cm}{Fluency} & \cellcolor{green!25} 5: Extremely fluent and natural. \\ 
 &  & \cellcolor{yellow!25} 4: Mostly fluent with minor awkwardness. \\ 
 &  & \cellcolor{yellow!50} 3: Moderately fluent with some awkwardness. \\ 
 &  & \cellcolor{orange!50} 2: Somewhat fluent with several awkward phrases. \\ 
 &  & \cellcolor{red!50} 1: Not fluent and very awkward. \\ \cline{2-3} 
 & \multirow{5}{4cm}{Clarity} & \cellcolor{green!25} 5: Extremely clear and easy to understand. \\ 
 &  & \cellcolor{yellow!25} 4: Mostly clear with minor ambiguities. \\ 
 &  & \cellcolor{yellow!50} 3: Moderately clear with some ambiguities. \\ 
 &  & \cellcolor{orange!50} 2: Somewhat unclear with several ambiguities. \\ 
 &  & \cellcolor{red!50} 1: Very unclear and difficult understand. \\ \hline
\end{tabular}
}
\caption{Evaluation criteria we employ for Arabic writing assistance tasks in our work.}
\label{tab:matrices}
    \end{table*}

\begin{table*}[t!]
    \centering
    \normalsize 
    \begin{tabularx}{\textwidth}{X}
    \toprule
        \textbf{Online Sources} \\
    \midrule
        \url{http://www.al-musheer.com/play-735.html} \\
        \url{https://www.wuduh1.com/2023/09/constructive-style.html} \\
        \url{https://learning.aljazeera.net/ar/grammar} \\
        \url{https://loghate.com/} \\
        \url{https://www.twinkl.ae/teaching-wiki/hmzt-alwsl-whmzt-alqt} \\
    \bottomrule
    \end{tabularx}%
    \caption{List of some of the publicly available online sources used in collecting \textit{Gazelle} dataset.}
    \label{online_resources}
\end{table*}

\input{}

\begin{landscape}
\begin{table}[]
\centering 

\resizebox{\columnwidth}{!}{

\begin{tabular}{lllllllllll}
\hline
\multicolumn{1}{c}{\textbf{Task}} &
  \textbf{Accuracy} &
  \textbf{Clarity} &
  \textbf{Helpfulness} &
  \textbf{\begin{tabular}[c]{@{}l@{}}Appropriateness \\ of MWEs\end{tabular}} &
  \textbf{Relevance} &
  \textbf{\begin{tabular}[c]{@{}l@{}}Consistency with \\ Gr. Rules \\ and Definition\end{tabular}} &
  \textbf{\begin{tabular}[c]{@{}l@{}}Depth of \\ Explanation\end{tabular}} &
  \textbf{\begin{tabular}[c]{@{}l@{}}Correctness \\ of I’rab\end{tabular}} &
  \textbf{\begin{tabular}[c]{@{}l@{}}Quality of \\ Text Refinement\end{tabular}} &
  \textbf{Fluency} \\  \hline

\textbf{GEC Cohere} &
  2.80 &
  3.28 &
  3.25 &
  \multicolumn{1}{c}{} &
   &
   &
   &
   &
   &
   \\ 
\textbf{GEC GPT4} &
  2.02 &
  3.17 &
  2.90 &
  \multicolumn{1}{c}{} &
   &
   &
   &
   &
   &
   \\
\rowcolor[HTML]{CCFFCC} 
\textbf{GEC GPT4o} &
  \textbf{3.27} &
  \textbf{3.77} &
  \textbf{3.63} &
  \multicolumn{1}{c}{\cellcolor[HTML]{CCFFCC}\textbf{}} &
  \textbf{} &
  \textbf{} &
  \textbf{} &
  \textbf{} &
  \textbf{} &
  \textbf{} \\
\textbf{GEC Gemini15pro} &
  2.70 &
  3.18 &
  3.03 &
  \multicolumn{1}{c}{} &
   &
   &
   &
   &
   &
   \\ \hline
\textbf{MWEs Cohere} &
   &
  3.68 &
   &
  3.15 &
  3.78 &
   &
   &
   &
   &
   \\
\textbf{MWEs GPT4} &
   &
  3.25 &
   &
  2.63 &
  3.28 &
   &
   &
   &
   &
   \\
\rowcolor[HTML]{CCFFCC} 
\textbf{MWEs GPT4o} &
  \textbf{} &
  \textbf{3.89} &
  \textbf{} &
  \textbf{3.46} &
  \textbf{3.91} &
  \textbf{} &
  \textbf{} &
  \textbf{} &
  \textbf{} &
  \textbf{} \\
\textbf{MWEs Gemini15pro} &
   &
  3.38 &
   &
  2.91 &
  3.36 &
   &
   &
   &
   &
   \\ \hline
\textbf{Gr. Rules Cohere} &
   &
  4.20 &
   &
   &
   &
  3.74 &
  4.03 &
   &
   &
   \\
\textbf{Gr. Rules GPT4} &
   &
  3.23 &
   &
   &
   &
  2.95 &
  3.11 &
   &
   &
   \\
\rowcolor[HTML]{CCFFCC} 
\textbf{Gr. Rules GPT4o} &
  \textbf{} &
  \textbf{4.73} &
  \textbf{} &
  \textbf{} &
  \textbf{} &
  \textbf{4.58} &
  \textbf{4.66} &
  \textbf{} &
  \textbf{} &
  \textbf{} \\
\textbf{Gr. Rules Gemini15pro} &
   &
  4.22 &
   &
   &
   &
  4.01 &
  4.33 &
   &
   &
   \\ \hline
\textbf{I'rab Cohere} &
   &
  4.41 &
  4.40 &
   &
   &
   &
   &
  3.91 &
   &
   \\
\textbf{I'rab GPT4} &
   &
  3.59 &
  3.44 &
   &
   &
   &
   &
  3.01 &
   &
   \\
\rowcolor[HTML]{CCFFCC} 
\textbf{I'rab GPT4o} &
  \textbf{} &
  \textbf{4.68} &
  \textbf{4.71} &
  \textbf{} &
  \textbf{} &
  \textbf{} &
  \textbf{} &
  \textbf{4.39} &
  \textbf{} &
  \textbf{} \\
\textbf{I'rab Gemini15pro} &
   &
  3.73 &
  3.83 &
   &
   &
   &
   &
  3.65 &
   &
   \\ \hline
\textbf{Text Cohere} &
   &
  3.48 &
   &
   &
   &
   &
   &
   &
  3.20 &
  3.39 \\
\textbf{Text GPT4} &
   &
  3.54 &
   &
   &
   &
   &
   &
   &
  3.15 &
  3.49 \\
\rowcolor[HTML]{CCFFCC} 
\textbf{Text GPT4o} &
  \textbf{} &
  \textbf{3.91} &
  \textbf{} &
  \textbf{} &
  \textbf{} &
  \textbf{} &
  \textbf{} &
  \textbf{} &
  \textbf{3.55} &
  \textbf{3.70} \\
\textbf{Text Gemini15pro} &
   &
  3.47 &
   &
   &
   &
   &
   &
   &
  3.39 &
  3.47 \\ \hline
\end{tabular}
}
\caption{Evaluation results for the five subtasks across all LLMs.}
\label{tab:stats evaluation}
\end{table}
\end{landscape}

\newpage

\end{document}